\documentclass{article} 
\usepackage[preprint]{colm2025_conference}

\usepackage{microtype}
\usepackage{hyperref}
\usepackage{url}
\usepackage{booktabs}

\usepackage{lineno}

\definecolor{darkblue}{rgb}{0, 0, 0.5}
\hypersetup{colorlinks=true, citecolor=darkblue, linkcolor=darkblue, urlcolor=darkblue}

\usepackage{amsmath}
\usepackage{amssymb}
\usepackage{mathtools}
\usepackage{amsthm}

\usepackage[capitalize,noabbrev]{cleveref}

\usepackage{enumitem}

\usepackage{amsmath,amsfonts,bm}

\def\eqref#1{equation~\ref{#1}}

\def\1{\bm{1}}

\def\vy{{\bm{y}}}

\DeclareMathAlphabet{\mathsfit}{\encodingdefault}{\sfdefault}{m}{sl}
\SetMathAlphabet{\mathsfit}{bold}{\encodingdefault}{\sfdefault}{bx}{n}

\usepackage[utf8]{inputenc} 
\usepackage[T1]{fontenc}    
\usepackage{hyperref}       
\usepackage{url}            
\usepackage{amsfonts}       
\usepackage{nicefrac}       
\usepackage{microtype}      
\usepackage{enumitem}

\usepackage{hyperref}
\usepackage{url}
\usepackage{multirow}
\usepackage{graphicx}
\usepackage{siunitx}  
\usepackage{tabularx}
\usepackage{arydshln}
\usepackage[flushleft]{threeparttable}
\usepackage{multirow}

\usepackage{enumitem}
\usepackage{amsmath}
\usepackage{xspace}
\usepackage{cleveref}
\usepackage{caption}
\usepackage{subcaption}
\usepackage{pgfplots}
\usepackage{tikz}
\usepackage{wrapfig}
\usepackage{fdsymbol}
\usepackage{listings}
\usepackage{float}

\usepackage{adjustbox}
\usepackage{tikz}
\usepackage{pgfplots}
\pgfplotsset{compat=1.18}
\usepackage{comment}
\usepackage{enumitem}

\usepackage{tcolorbox}  
\usepackage{listings}   

\usepackage{fontawesome5}

\newtcolorbox{textbox}[1][]{
  breakable,
  colframe=black,          
  colback=white,            
  coltitle=white,           
  title=#1,                 
  rounded corners,          
  boxrule=0.5mm,            
  boxsep=5pt,               
  toptitle=1mm,             
  bottomtitle=1mm,          
  left=10pt,                
  right=10pt,               
  top=2pt,                  
  bottom=2pt,               
  fonttitle=\bfseries       
}

\theoremstyle{plain}

\theoremstyle{definition}

\theoremstyle{remark}

\definecolor{lowyellow}{RGB}{241, 196, 15}

\definecolor{earthyellow}{RGB}{225, 169, 95}

\title{StructTest: Benchmarking LLMs' Reasoning through \\ Compositional Structured Outputs}

\author{{Hailin Chen}$^*$$^2$
\And
\textbf{Fangkai Jiao}$^*$$^1$$^,$$^3$
\And
\textbf{Mathieu Ravaut}$^*$$^1$$^,$$^3$
\And
\textbf{Nawshad Farruque}$^*$$^4$
\And
\textbf{Xuan Phi Nguyen}$^*$$^2$
\And
\textbf{Chengwei Qin}$^1$
\And
\textbf{Manan Dey}$^2$
\And
\textbf{Bosheng Ding}$^1$
\AND
{Caiming Xiong}$^2$
\quad
{Shafiq Joty}$^{1,2\dagger}$
\quad
\textbf{Yingbo Zhou}$^{2\dagger}$\\
\\
$^1$ Nanyang Technological University, Singapore
\quad
$^2$ Salesforce Research \\
$^3$ Institute of Infocomm Research (I$^{2}$R), A$^{*}$STAR, Singapore \\
$^4$ Dept. of Computing Science, University of Alberta, Canada
}

\begin{document}

\ifcolmsubmission
\linenumbers
\fi

\def\thefootnote{*}\footnotetext{Equal contribution. Correspondence to: \texttt{\small{hailin001@ntu.edu.sg}}}\def\thefootnote{\arabic{footnote}}
\def\thefootnote{\dagger}\footnotetext{Equal mentorship}\def\thefootnote{\arabic{footnote}}

\maketitle


\vspace{-2em}
\begin{center}
\faIcon{github} Code \& Data: \href{https://github.com/SparkJiao/StructTest}{https://github.com/SparkJiao/StructTest}
\end{center}

\begin{abstract}


The rapid advancement of large language models (LLMs) demands robust, unbiased, and scalable evaluation methods. However, human annotations are costly to scale, model-based evaluations are susceptible to stylistic biases, and target-answer-based benchmarks are vulnerable to data contamination and cheating. To address these limitations, we propose \textbf{StructTest}, a novel benchmark that evaluates LLMs on their ability to {follow compositional instructions and generate structured outputs}, providing an unbiased, cost-effective, and difficult-to-cheat evaluation framework. Assessments are conducted deterministically using a rule-based evaluator, which can be easily extended to new tasks {and datasets}. By testing structured outputs across diverse domains—including Summarization, Code, HTML, and Math—{and evaluating 17 popular LLMs, we demonstrate that StructTest remains challenging even for top-performing models like Deepseek-V3/R1 and GPT-4o, establishing it as a robust proxy for measuring reasoning capabilities.} We believe StructTest offers a critical and complementary approach to achieving objective and comprehensive model evaluation.



\end{abstract}


\section{Introduction}

\begin{figure}[ht]
    \centering
    \begin{subfigure}[b]{0.35\textwidth}
        \includegraphics[width=\linewidth]{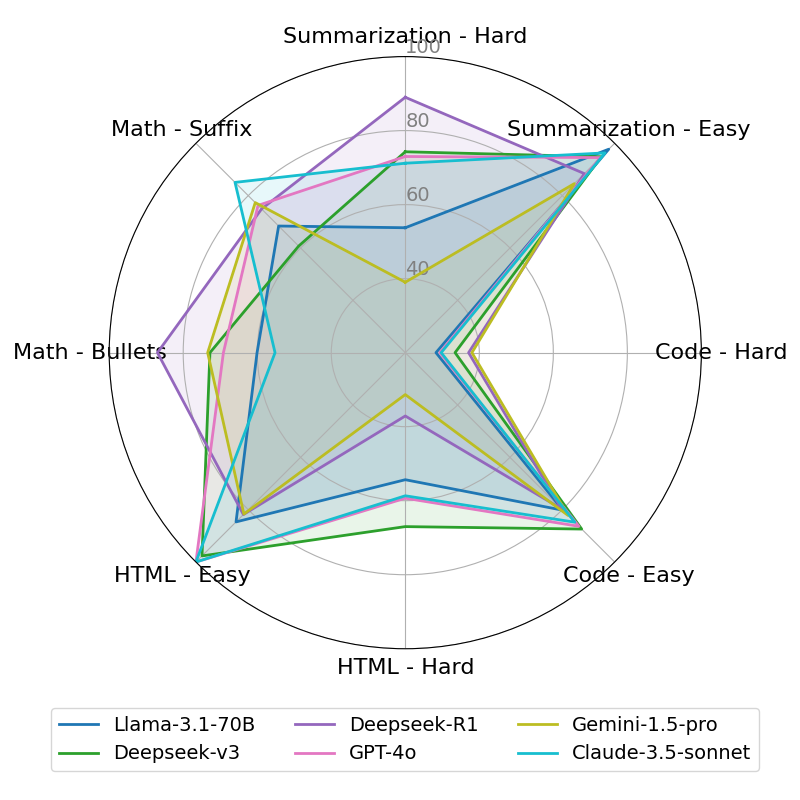}
        \caption{\small Comparison of top models on StructTest.  DeepSeek-v3 and R1 consistently outperform others, achieving the highest scores across nearly all challenging benchmarks.}
        \label{fig:compare}
    \end{subfigure}
    \hfill
    \begin{subfigure}[b]{0.62\textwidth}
        \includegraphics[width=\linewidth]{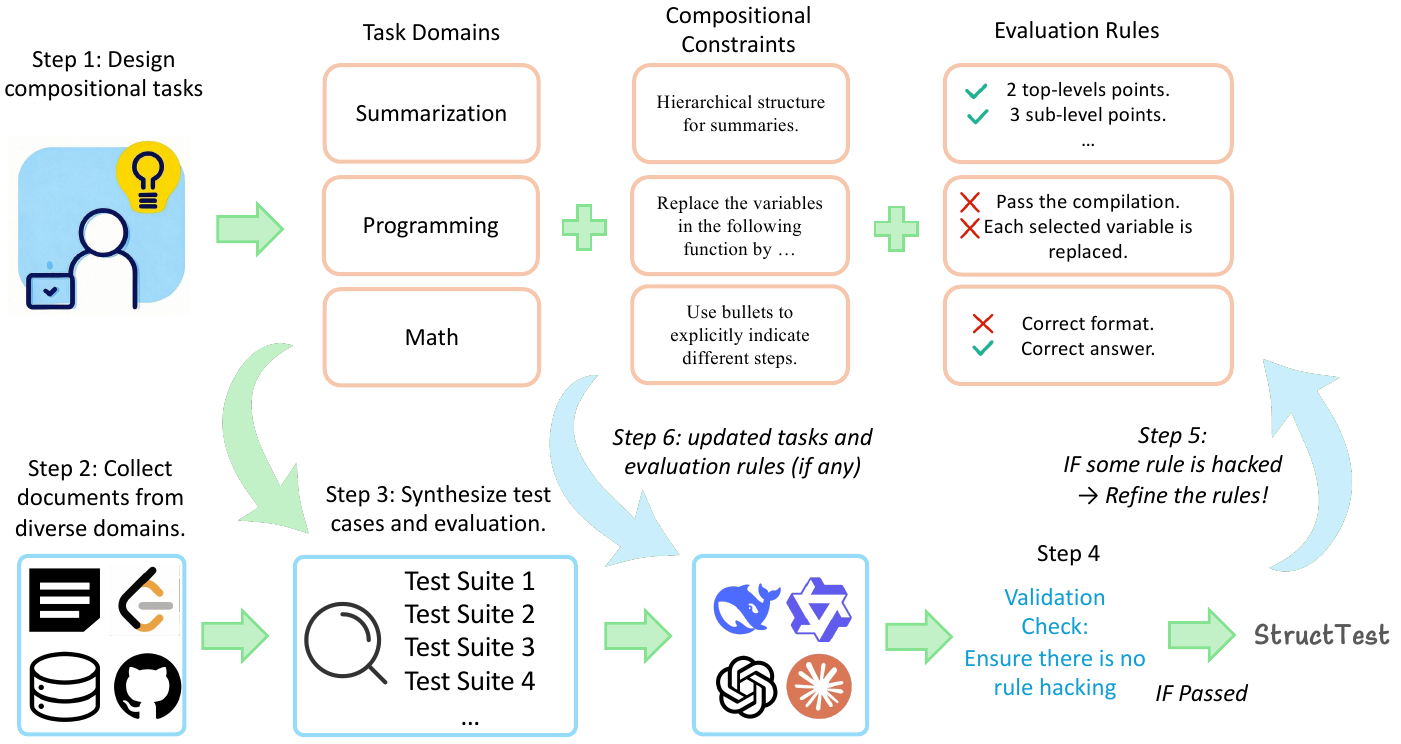}
        \caption{\small The construction process of StructTest highlights its adaptability to varying difficulty levels, extendability, and scalability. Tasks are designed independently of underlying data, and evaluation uses rule-based methods that can be dynamically updated to address uncaptured cases. This flexibility ensures StructTest remains robust and relevant over time.}
        \label{fig:framework}
    \end{subfigure}
    \label{fig:total}
\end{figure}



Since the launch of ChatGPT two years ago, we have witnessed an unprecedented surge in the development and release of large language models (LLMs). In 2024 alone, major tech companies, including OpenAI, Anthropic, Google, and xAI, have rolled out new iterations of their flagship proprietary models. Meanwhile, the open-source community has been even more active, producing a substantially larger number of models. To showcase advancements, many of these models have relied on established benchmarks like MMLU \citep{hendrycks2020measuring} and GSM8K \citep{DBLP:journals/corr/abs-2110-14168}. However, as we will explore in \cref{sec:related_work}, current benchmarks face significant limitations: (1) human annotations are costly to obtain, maintain, and scale; (2) model-based evaluations are heavily influenced by model bias; and (3) target-answer-based datasets are vulnerable to data contamination. Consequently, there is an urgent need for a new evaluation benchmark that is (1) cost-effective and easy to implement, (2) free from bias, and (3) resilient to data contamination.



To address these challenges, we propose \textbf{StructTest}, a benchmark designed to evaluate compositional instruction-following capabilities through structured outputs. In StructTest, models are required to generate outputs conforming to a predefined structure. As illustrated in Figure~\ref{fig:framework}, this structure is defined as $\langle\,\mathrm{Domain\ Task},\;\mathrm{Format\ Rules}\,\rangle$, where the Domain Task specifies the semantic or reasoning objective (e.g., summarization or program execution simulation), and the Format Rules dictate the precise output structure (e.g., hierarchical summarization with a fixed number of key points, or JSON-formatted outputs with program-specified keys). The model responses are evaluated for both structural accuracy and, where relevant, the correctness of the parsed content. Reasoning capabilities are also assessed, as the problem-solving process in StructTest requires critical abilities such as decomposing instructions, understanding and retaining subtle constraints during extended decoding, and executing logical actions—skills closely aligned with complex reasoning~\citep{li2025codeio} and agentic workflows \citep{weng_blog_agent}. 

Crucially, StructTest employs programmatic evaluation, ensuring that the assessment is unbiased, efficient, and cost-effective. StructTest tasks are inherently compositional, allowing for adjustable difficulty levels. This design ensures that the benchmark remains relevant and challenging for future generations of LLMs. Furthermore, StructTest is decoupled from underlying task data, enabling easy sampling of new test sets and extension to novel tasks. This flexibility makes it robust to data contamination as we describe in \Cref{sec:robustness_scalability}. 

StructTest spans multiple task domains, including text summarization, code, HTML, and math. As illustrated in Figure~\ref{fig:compare}, evaluations on 17 popular LLMs reveal that even the most advanced models (e.g., DeepSeek-v3 and R1) show significant room for improvement, with many struggling on challenging domains such as Code-Hard and HTML-Hard. The reliance on underlying data from existing benchmarks in some subtasks, coupled with notable performance drops, raises concerns about data memorization rather than genuine generalization in current LLMs. Notably, StructTest achieves a Pearson correlation of over 92\% with both the human-annotated ChatBot Arena~\citep{chiang2024chatbotarenaopenplatform} and the widely used MMLU dataset, validating its effectiveness as a reliable proxy for general reasoning ability. Its design also ensures strong extensibility and robustness against data contamination.



\section{Literature Review} 
\label{sec:related_work}
The evaluation of LLMs has emerged as a crucial research focus, especially as these models are applied to diverse tasks demanding structured reasoning. Current evaluation methods can be categorized into three types: human-based, model-based, and target-answer-based. While each provides valuable insights, they also come with significant limitations.

\paragraph{Human-Based Evaluation Benchmarks}

A prominent example of human-based evaluation is \textbf{Chatbot Arena} \citep{chiang2024chatbotarenaopenplatform}, which uses human voting to calculate model ELO scores. While it provides reliable assessments, it faces significant limitations: high resource costs due to extensive human annotations, limited scalability to only a few models, and challenges in sustaining community engagement for evaluating the latest models.


\paragraph{Model-Based Evaluation Benchmarks}

Model-based evaluation frameworks leverage LLM-as-a-judge to assess the capabilities of other models. Notable examples include \textbf{MT-Bench} \citep{zheng2023judging}, \textbf{AlpacaEval} \citep{dubois2024lengthcontrolledalpacaevalsimpleway}, \textbf{Arena-Hard-Auto} \citep{li2024crowdsourceddatahighqualitybenchmarks}, and \textbf{Fofo} \citep{xia-etal-2024-fofo}. While these frameworks offer flexibility in evaluating diverse tasks, they are prone to biases. Most notable biases include: (i) \textbf{Cheating by Null-Models}: As noted by \citet{zheng2024cheatingautomaticllmbenchmarks}, even a null-model with constant responses can achieve high rankings in these benchmarks, raising significant concerns about their reliability. (2) \textbf{Length Bias}: \citet{dubois2024lengthcontrolledalpacaevalsimpleway} highlight that popular benchmark AlpacaEval favors responses of longer lengths, and shows the benefits of controlling for this factor. (3) \textbf{Positional Bias}: It has been demonstrated that the order in which responses are presented to LLM judges can influence their decisions in pairwise judgment tasks \citep{wang2023large} 
Moreover, \citet{park2024offsetbias} identify four additional biases that can affect LLM judges.
\paragraph{Target-Answer-Based Evaluation Benchmarks}

Target-answer-based evaluations assess model capabilities by comparing directly with reference answers. Most conventional LLM benchmarks fall into this category, including \textbf{ARC} \citep{DBLP:journals/corr/abs-1803-05457}, \textbf{GSM8K} \citep{DBLP:journals/corr/abs-2110-14168}, \textbf{BIG-Bench} \citep{DBLP:conf/naacl/ZhongCGLLWSCD24}, \textbf{AGIEval} \citep{DBLP:conf/naacl/ZhongCGLLWSCD24} and \textbf{MMLU} \citep{hendrycks2020measuring}. For instance, MMLU evaluates LLMs' reasoning abilities using curated datasets from various competitive exams. While these benchmarks are unbiased, they face a significant limitation: data contamination. The extensive use of internet-sourced datasets in pre-training LLMs often overlaps with benchmark datasets, leading to inflated performance metrics and compromising the validity of evaluations \citep{ravaut2024much}.

To address the limitations of existing evaluation benchmarks, we introduce StructTest, which evaluates structured generation as a proxy for general reasoning. While prior studies have explored how format instructions influence task performance \citep{he2024doespromptformattingimpact, DBLP:journals/corr/abs-2408-08656}, StructTest goes beyond simple formatting by incorporating compositional structured outputs across diverse domains. It is the first benchmark specifically designed to assess the general instruction-following capabilities of LLMs through structured outputs.

\section{StructTest Benchmark}



The primary goal of StructTest is to assess LLMs' ability to follow complex instructions for generating structured outputs that are decoupled from underlying data (minimizing contamination risks) and programmatically verifiable. This section details the tasks comprising the StructTest benchmark: summarization, coding, HTML generation, and mathematical reasoning. Many of these tasks demand reasoning capabilities, and all are designed to be adaptable to varying difficulty levels, extendable, and scalable.


\subsection{Summarization} 

As the first task in StructTest, we focus on summarization, a well-established domain for evaluating LLMs. Most existing research emphasizes the content of summaries, assessing aspects such as coherence \citep{chang2023booookscore}, faithfulness to the source \citep{laban-etal-2023-summedits}, coverage of diverse information \citep{huang2023embrace}, positional bias in context utilization \citep{ravaut-etal-2024-context}, and hallucination \cite{wan2024positional}. As LLMs advance, addressing complex user requirements for summaries becomes increasingly critical. For instance, \citet{liu2023benchmarking} benchmark LLMs on content-specific instructions. However, an equally important yet underexplored aspect is the style or format of summaries. To address this gap, we introduce three format-following tasks in summarization:


\paragraph{$\bullet~$ Length} Controlling summary length has been extensively researched \citep{liu-etal-2018-controlling,liu-etal-2022-length}. Users looking for more granular details will prompt the system to output longer summaries. 
To measure length-following ability, we verify whether the LLM's output $\vy$ contains the required number of sentences $N$, which is sampled uniformly from a fixed interval across data points. Formally:
\begin{equation}
\text{Score} =
\begin{cases} 
1, & \text{if } \text{len}(\vy) = N, \\
0, & \text{otherwise.}
\end{cases}
\end{equation}

\paragraph{$\bullet~$ Bullet points}

Bullet points are a natural method to summarize and have yielded several of the most widely used datasets in summarization research  \citep{hermann2015teaching,mukherjee-etal-2022-ectsum}. 
This format is appealing to users who wish to see a clear separation of ideas in the output summary. 
We prompt the LLM to summarize through a list of either unnumbered bullet (or other symbol) points, or numbered points, with a varying number of points (sampled uniformly from a fixed interval).

For unnumbered points, we check whether the output contains the specified symbol $S$ in the correct number of times $N$:
\begin{equation}
    \text{Score} =
        \begin{cases} 
        1, & \text{if } \text{count}(S \in \vy) = N, \\
        0, & \text{otherwise.}
    \end{cases}
\end{equation}
For numbered points, we verify that output lines ($\vy_{i}$, .., $\vy_{M}$) are of the appropriate count and start with the correctly ordered sequence of numbers:
\begin{equation}
    \text{Score} =
        \begin{cases} 
        1,& \text{if } (M=N) \land (\forall i \in [1, N], \vy_{i,0}=\text{str}(i)) \\
        0,& \text{otherwise.}
    \end{cases}    
\end{equation}



\paragraph{$\bullet~$ Questions}

Yet another approach to summarization consists in answering key questions about the source, most notably the 5 Wh-questions of what/why/who/when/where. Question-answering is a popular paradigm in summarization evaluation \citep{deutsch2021towards,scialom2021questeval,fabbri2021qafacteval}, as it naturally enables to review that key facts from the source are covered. 
To induce format following, we prompt the LLM to structure its summary such that it is composed of the list of 5 Wh-questions, each followed by its corresponding answer. This process is akin to query-focused summarization \citep{vig-etal-2022-exploring}, where the Wh-questions form the query. 

To evaluate Wh-questions summary formatting, we check that summary lines start with the Wh-questions. We also enforce that all questions are present, in any order. Formally, following the previous notation and noting $\bm{Q}$ the set of Wh-questions:
\begin{equation}
    \text{Score} =
        \begin{cases} 
        1,& \text{if } (\bm{Q} \subset \vy) \land (\forall i \in [1, N], \vy_{i,0} \in \bm{Q}) \\
        0,& \text{otherwise.}
    \end{cases}
\end{equation}


\paragraph{Creating more complex tasks}

StructTest instructions following one of the aforementioned summarization formats are referred to as \textbf{Easy Summarization}. To evaluate the compositional reasoning capability of LLMs, we compose different format instructions together.
We use two types of combinations: first, we specify the number of bullet points or numbered points and the desired length (in sentences) of each point; second, we ask the LLM to nest bullet points within existing points, where nested points start with the tab symbol marking indentation. These instructions combining two summarization formats are referred to as \textbf{Hard Summarization}. In this latter case, evaluation metrics defined above are also combined together and the LLM needs to verify each property. We provide examples for each task in \Cref{appendix:task_examples}.

\subsection{Code} 




Programming languages, with their structured and rule-based nature, provide an ideal framework for evaluating LLMs' instruction-following capabilities. Compilers and interpreters efficiently validate correctness, offering clear binary feedback on whether generated code meets syntactic and functional requirements. This makes coding tasks a practical and scalable benchmark for assessing LLM performance in structured environments.


Given the widespread use of programming in daily workflows, and the significant success and adoption of Code-LLMs in real-world systems \citep{jimenez2024swebench,xie2024osworld}, understanding complex instructions in a language code-interleaved environment has become increasingly important. 
To closely measure the capability of LLMs in application-like scenarios, we have developed the following tasks:


\paragraph{$\bullet~$ Add `{print}' statements}

A common code editing task involves revising a code snippet. We propose a simple edit task where the LLM is asked to add a `{print}' statement \textbf{each time a new variable is initialized.} We create two datasets based on difficulty: the \textbf{Easy} set contains code snippets with 3 to 30 lines, while the \textbf{Hard} set includes snippets with 50 to 200 lines.

Since the instruction is fixed, we can programmatically generate the expected code snippet. Specifically, we use the ast package\footnote{\href{https://docs.python.org/3/library/ast.html}{https://docs.python.org/3/library/ast.html}} to parse the abstract syntax tree and extract variable initializations. The expected target code is then synthesized by inserting print statements using predefined templates. The evaluation metric is \textbf{exact match}, comparing the predicted code snippet with the synthesized one.



\paragraph{$\bullet~$ Replace variables}

Another edit-based task involves \emph{replacing variables}. For data construction, we first use the ast package to extract variables from a code snippet and randomly generate meaningless strings as target variable names. We then create a mapping from the original variable names to the target ones and include this mapping in the instruction, asking the LLM to replace all instances of the source variables with the corresponding target variables. The expected code snippet is generated by performing string replacements according to the mapping. The LLM's output is evaluated using \textbf{exact match}, comparing its prediction with the synthesized expected program.



\paragraph{$\bullet~$ Test case input generation}

As a fundamental aspect of software engineering, writing high-quality unit tests (i.e., sample input-output pairs) is crucial for verifying program correctness. Given that predicting unit test outputs remains challenging for current LLMs \citep{li2022alphacode,jain2024livecodebench,jiao2024pfpo}, we simplify the task by asking LLMs to generate 5 distinct groups of test case inputs for a given programming problem and its corresponding solution. Successfully generating these inputs requires the model to comprehend both the problem and the program, making reasoning an essential requirement.

We evaluate the validity by executing the program on the {predicted test case inputs}, and if no runtime error is raised for all inputs, the generation is deemed correct. We use the \textbf{averaged pass rate} over all problems as the evaluation metric.


\paragraph{$\bullet~$ Simulate program execution}



Simulating program execution poses several challenges, including understanding and following each action in the program, tracking runtime variable states, and associating them to produce the correct outcome. These tasks are closely tied to reasoning and agent-based operations, making program simulation a valuable proxy for assessing the ability to follow compositional instructions and perform logical reasoning. We prompt the LLM to simulate the step-by-step execution of a given program with specific inputs and derive the expected output. The task is divided into two difficulty levels—\textbf{Easy} and \textbf{Hard}—based on the length of the code snippet being simulated.


For the Easy level, we include multiple test cases from the original dataset for each question to ensure robust evaluation. If all predicted outputs \textbf{exactly match} the ground-truth ones, the generation for the question is considered as correct. For the Hard level, we evaluate using only one simple test case, as (1) the complexity of the code snippets themselves is sufficiently challenging, and (2) scaling test cases uniformly is difficult—some may involve millions of input numbers in a single line. The final metric is the average \textbf{exact match} rate across all questions.




\subsection{HTML Generation} \label{subsec:Structured-HTML}
The use of LLMs in generating websites has been recognized as a valuable task, reducing the workload for web designers and developers while democratizing web development for non-technical users \cite{calo2023leveraging}. In these applications, adhering to user-specified HTML structures is critical. \citet{tang2023struc} highlight that LLMs often struggle to generate structured HTML, though their study is limited to simple structures and relies on content-based evaluation requiring human assessment. 


In contrast, we formulate this task as to generate a specific number of standard HTML tags (``html'', ``head'', ``title'', ``div'', ``body'', ``h1'', ``h2'', ``p'',``footer'') as instructed  with the following structural constraints: ``title'' should be nested inside ``head'', ``div'' and ``footer''  are nested inside ``body'', and the rest of the tags are nested inside ``div''. An example prompt with our prompt template is:\footnote{The number of each tag except ``html'' varies across examples. For ``html'', it is fixed to 1.} \emph{``Generate only an html code that has 1 html tag. Inside the html tag, generate 1 head tag and 1 body tag. Inside of each head tag, generate 1 title tag and inside of each body tag, generate 2 div tags and 1 footer tag. Inside of each div tag, generate 1 h1 tag, 1 h2 tag and 1 p tag. Your generated html code: \footnote{We provide the full prompt in the Appendix \ref{appendix:task_examples} Figure \ref{fig:exp-html-gen}. In the full prompt we also provide a fewshot example and instructions regarding limiting the generation to only the html code block.}''}, 
and the expected generation should be an html code block as the right one: 

\begin{wrapfigure}[9]{r}{0.4\textwidth} 
\centering
\vspace{-0.7cm}
\begin{adjustbox}
{width=0.96\linewidth,height=1.9cm,keepaspectratio}
\begin{lstlisting}[language=HTML, basicstyle=\footnotesize]
<html>
    <head>
        <title></title>
    </head>
    <body>
        <div>
            <h1></h1>
            <h2></h2>
            <p></p>
        </div>
        <div>
            <h1></h1>
            <h2></h2>
            <p></p>
        </div>
        <footer></footer>
    </body> 
</html>
\end{lstlisting}
\end{adjustbox}
\end{wrapfigure}
The counts of each tag to be generated are sampled uniformly from a fixed interval. Based on the range of the interval, we create two sets, \textbf{Easy} where the interval range is 2-5, and \textbf{Hard} where the range is 2-12. Successful completion of this task requires decomposition of the instruction and retraining constraints, which are important for reasoning.



We consider a generation to be successful if the count of the tags is equal to the ones provided in the prompt taking into account their nested structure and all the tags are properly formatted, i.e., an opened HTML tag has to be closed.

\subsection{Math Reasoning} 

Mathematical reasoning is a common task in LLM evaluations, with benchmarks like GSM8K and MATH \citep{llm-eval-harness,gsm8k_cobbe2021training,math_dataset_hendrycks2021measuring}. However, the influence of varying format templates on these tasks is often overlooked, potentially leading to inconsistencies, as many studies may not use impartial templates \citep{metamath_yu2023metamath,deepseekmath_shao2024,chain_of_thought_wei2022chain,toshniwal2024openmathinstruct}. The variability in solutions—ranging from numbers and fractions to LaTeX expressions—means extraction methods may differ across studies, resulting in biased comparisons that favor models optimized for specific frameworks. For instance, MetaMathQA \citep{metamath_yu2023metamath} created a dataset where answers follow specific phrases which their evaluation procedure uses to extract answers, disadvantaging models that do not adhere to these phrases.

A reliable model should not only produce correct answers but also consistently present a chain of thought (CoT) in a predefined format \citep{chain_of_thought_wei2022chain}. Reliably extracting reasoning steps can be advantageous, such as for generating thought chains for process supervision \citep{prm_lightman2023let}. Therefore, we structure our math evaluations around two key aspects: final answer parsing and chain of thought bullet point formatting.

\paragraph{$\bullet~$ Final answer parsing} 

We designed 7 distinct styles for final answer presentation and created prompts to instruct models to follow these styles. Python rules were implemented to assess whether a model's response adheres to the assigned style. For evaluation, we used standard benchmarks like GSM8K, assigning each question a random style. This allows us to measure format consistency accuracy, which, when combined with math accuracy, provides a fairer and more comprehensive comparison across LLMs. Specifically, only answers that are both mathematically correct and format-compliant are considered correct. In our setup, final answer parsing is categorized as \textbf{Easy}.


\paragraph{$\bullet~$ CoT bullet points.}

Solutions to math problems often involve multiple reasoning steps, and we designed 5 distinct presentation styles. These include Markdown formats like ``**Step 1** ...'', and JSON structures. We also defined a range for the number of steps, requiring models to adjust step granularity. For simpler solutions, models should break down steps into finer details, while for complex solutions, they should consolidate multiple steps into longer ones. By pairing each bullet point style with a unique final answer style, we created 20 formats, classified as \textbf{Hard}. We hypothesize that some LLMs may find these styles intuitive, while others may struggle, potentially leading to significant performance variations, as discussed in Section \ref{section:results}. Although the number of styles could theoretically be infinite, we rely on manually crafted styles to ensure accuracy and consistency.


\subsection{Robustness to Contamination and Benchmark Scalability} \label{sec:robustness_scalability}

A key challenge in benchmarking LLMs is the risk of data contamination, where models are exposed to test data during training. StructTest addresses this issue in two ways. First, its tasks are designed to be highly unlikely to have been encountered by existing models during training. By focusing on carefully constructed structured output tasks, StructTest significantly reduces the risk of data contamination. Second, the benchmark's flexible design allows for periodic updates, including new StructTest samples, new underlying task data, additional task domains, and varying complexity levels. 
To future-proof the benchmark, we aim to maintain a confidential, held-out test set that is regularly updated, ensuring that model performance reflects true generalization capabilities rather than memorization of pre-exposed data.

StructTest is highly scalable, enabling seamless extension to new tasks through the creation of new prompts and rule-based evaluation methods. It supports cost-efficient evaluation for new models, requiring minimal overhead beyond inference costs. This design ensures flexibility and adaptability to meet diverse evaluation needs.





\section{Evaluation Results}


\subsection{Results Overview}
\label{section:results}

We evaluated the StructTest benchmark against a representative selection of open-source and closed-source models.\footnote{See \Cref{appendix:model_versions} for detailed versions of closed-source models} \Cref{table:04_main_results} summarizes the overall results across all domains of StructTest for all LLMs. For open-source models, we used their instruction-tuned versions rather than the base model. 
Notably, the top-performing LLM, {DeepSeek-R1}, achieves only 74.76\% on StructTest-All and 67.10\% on StructTest-Hard, underscoring the benchmark's high level of difficulty. Additionally, {GPT-4o} is a close runner-up, and closed-source models generally outperform their open-source counterparts.

\begin{table*}[t]
\centering
\resizebox{0.95\linewidth}{!}{
\begin{tabular}{lccc:cccccccc}
\toprule
\multicolumn{1}{l}{\multirow{2}{*}{\textbf{LLM}}} & \multicolumn{3}{c}{\textbf{Average}} & \multicolumn{2}{c}{\textbf{Summarization}} & \multicolumn{2}{c}{\textbf{Code}} & \multicolumn{2}{c}{\textbf{HTML}} & \multicolumn{2}{c}{\textbf{Math}}  \\
\cmidrule(lr){2-4}
\cmidrule(lr){5-6}
\cmidrule(lr){7-8}
\cmidrule(lr){9-10}
\cmidrule(lr){11-12}
\multicolumn{1}{c}{}                              & \textbf{All} & \textbf{Easy} & \textbf{Hard} & \textbf{Easy} & \textbf{Hard}       & \textbf{Easy}   & \textbf{Hard}   & \textbf{Easy}   & \textbf{Hard}   & \textbf{Easy}  & \textbf{Hard}                                \\
\toprule
DS-Dis-Qwen-1.5B                                  & 9.19            & 14.31           & 4.07            & 26.90                & 10.08               & 24.58           & 4.21            & 0.00              & 0.00              & 5.76              & {1.97} \\
Phi-3-mini-128k                                   & 19.97           & 32.65           & 7.30            & 55.83                & 11.39               & 51.56           & 13.33           & 0.00              & 0.00              & 23.20             & 4.47              \\
Qwen-2-7B                                         & 17.55           & 28.58           & 6.53            & 48.79                & 9.50                & 50.63           & 13.27           & 0.33              & 0.00              & 14.56             & 3.34              \\
Mistral-7B                                        & 13.82           & 21.83           & 5.81            & 51.29                & 15.89               & 31.25           & 5.96            & 1.67              & 0.33              & 3.11              & 1.06              \\
Llama-3.1-8B                                      & 37.20           & 46.05           & 28.35           & 87.23                & 52.50               & 50.42           & 17.29           & 12.00             & 0.33              & 34.57             & 43.29             \\
Mistral-nemo                                      & 27.20           & 43.76           & 10.64           & 72.83                & 18.36               & 63.13           & 17.81           & 6.33              & 0.00              & 32.75             & 6.37              \\
Mixtral-8x7B                                      & 17.73           & 28.93           & 6.54            & 67.38                & 16.78               & 33.33           & 3.97            & 3.33              & 0.33              & 11.68             & 5.08              \\
Llama-3.1-70B                                     & 65.93           &\underline{82.75}& 49.10           & \textbf{97.73}       & 53.75               & 80.21           & 28.29           & 84.67            & 54.33             & \underline{68.39} & 60.05             \\
DeepSeek-v3                                       &\underline{73.52}& \textbf{85.16}  &\underline{61.88}& \underline{94.85}    & \underline{74.28}   & \textbf{87.40}  &\underline{33.45}& \textbf{97.67}    & \textbf{67.00}    & 60.73             & \underline{72.78}   \\
DeepSeek-R1                                       &\textbf{74.76}   & 82.42           & \textbf{67.10}  & 88.42                & \textbf{89.00}      &\underline{81.85}& \textbf{37.11}  & \underline{86.67} & \underline{55.33}         & \textbf{74.91}    & \textbf{86.96}    \\
\hdashline GPT-3.5-turbo                          & 38.27           & 61.75           & 14.79           & 86.35                & 22.33               & 74.48           & 19.38           & 47.67             & 6.00              & 38.51             & 11.45             \\
GPT-4o-mini                                       & 60.04           & 75.93           & 44.16           & \textbf{98.83}       & \textbf{75.58}   & 82.40           & 25.67           & 45.00             & 7.67              & 77.48             & 67.70             \\
GPT-4o                                            & \textbf{73.16}  & 89.04 & \textbf{57.29}  & 94.54                & \underline{73.00}      & \textbf{86.36}  & 29.34           & \underline{99.00} & \underline{57.67} & 76.27             & \underline{69.14}   \\
Gemini-1.5-pro                                    & 63.44           & 81.44           & 45.44           & 84.58                & 39.03               & 82.19           & \textbf{38.01}  & 81.67             & 31.33             & 77.33             & \textbf{73.39}    \\
Claude-3-haiku                                    & 36.15           & 53.31           & 18.99           & 72.19                & 22.06               & 66.25           & 22.18           & 41.00             & 10.33             & 33.81             & 21.38             \\
Claude-3-opus                                     & 68.81           & \underline{89.14}           & 48.48 & 91.21                & 46.19               &\underline{85.00} & \underline{36.04} & \textbf{100.00}   & 56.67             & \underline{80.36} & 55.04             \\
Claude-3.5-sonnet                                 &\underline{72.62} & \textbf{91.55}  & \underline{53.69}           & \underline{96.33}    & 71.19               & 84.79           & 29.70           & \textbf{100.00}   & \textbf{58.67}    & \textbf{85.06}    & 55.19             \\ 
\bottomrule
\end{tabular}
}
\caption{Overview of Evaluation Results on StructTest. {The best results under each task} are highlighted in \textbf{bold}, and the second-best are \underline{underlined}. DS-Dis-Qwen-1.5B refers to DeepSeek-R1-Distill-Qwen-1.5B. {DeepSeek-R1 demonstrates strong performance, even surpassing some closed-source models on the Hard split. However, there remains significant room for improvement, highlighting the challenging nature of StructTest.}}
\label{table:04_main_results}
\end{table*}


\paragraph{Summarization Results}

As we see in \Cref{table:04_main_results}, overall, on summarization tasks, the Llama-3.1 and DeepSeek series stands out among open-source models, delivering performance comparable to GPT-4 on the Easy subset (e.g., 97.73 for Llama-3.1-70B vs. 98.83 for GPT-4o-mini). On average, closed-source LLMs outperform open-source models, especially on the Hard subset. The exception is DeepSeek-v3 and R1, which excel even beyond closed-source models on the Hard split.

Open-source LLMs other than DeepSeek experience a 70\% drop in accuracy on Hard tasks compared to Easy ones, while closed-source models show a 55\% relative decline. This performance gap showcases the significant challenge that LLMs face in adhering to more complex formatting instructions.


\begin{table}[t]
\centering
\setlength{\tabcolsep}{2.0pt}
\resizebox{0.9\linewidth}{!}{
\begin{tabular}{lccccccc}
\toprule
\multirow{3}{*}{\textbf{LLM}} 
& \multicolumn{4}{c}{\textbf{Easy}} 
& \multicolumn{3}{c}{\textbf{Hard}} \\
\cmidrule(lr){2-5}
\cmidrule(lr){6-8}
& \textbf{Length} 
& \textbf{\begin{tabular}[c]{@{}c@{}}Bullet\\ points\end{tabular}} 
& \textbf{\begin{tabular}[c]{@{}c@{}}Numbered\\ points\end{tabular}} 
& \textbf{\begin{tabular}[c]{@{}c@{}}Wh-\\ questions\end{tabular}} 
& \textbf{\begin{tabular}[c]{@{}c@{}}Bullets \\ + length\end{tabular}} 
& \textbf{\begin{tabular}[c]{@{}c@{}}Numbers \\ + length\end{tabular}} 
& \textbf{\begin{tabular}[c]{@{}c@{}}Indented\\ points\end{tabular}} \\
\midrule 

{DS-Dis-Qwen-1.5B}                             & 1.00 & 9.33 & 87.92 & 9.33 & 8.25 & 21.75 & 0.25 \\
Phi-3-mini-128k                              & 32.25 & 30.25 & 88.08 & 72.75 & 9.58 & 23.92 & 0.67 \\
Qwen-2-7B                                    & 27.25 & 68.25 & \underline{99.67} & 0.00 & 20.00 & 8.42 & 0.08 \\ 
Mistral-7B                                   & 27.67 & 57.50 & \underline{99.67} & 20.33 & 19.75 & 27.50 & 0.42 \\
Llama-3.1-8B                                 & \underline{93.83} & \underline{99.67} & 99.00 & 56.42 & 62.25 & 61.75 & 33.50 \\
Mistral-nemo                                 & 53.25 & 94.58 & 98.92 & 44.58 & 26.08 & 27.42 & 1.58 \\
Mixtral-8x7B                                 & 33.33 & 72.50 & 86.25 & 77.42 & 19.17 & 30.25 & 0.92 \\
Llama-3.1-70B                                & \textbf{94.75} & \textbf{99.92} & \textbf{99.92} & \underline{96.33} & 64.17 & \underline{64.83} & 32.25 \\
DeepSeek-v3                                  & 81.33 & 98.92 & 99.17 & \textbf{100.00} & \underline{72.58} & 73.08 & \underline{77.17} \\ 
DeepSeek-R1                                  & 83.08 & 94.50 & 99.17 & 76.92 & \textbf{89.33} & \textbf{92.08} & \textbf{85.58} \\ 
\hdashline
GPT-3.5-turbo                                & 48.42 & 99.67 & \textbf{99.92} & 97.42 & 26.08 & 32.58 & 8.33 \\
GPT-4o-mini                                  & \textbf{97.25} & \underline{99.92} & \textbf{99.92} & \underline{98.25} & \textbf{75.33} & \textbf{76.83} & \underline{74.58} \\ 
GPT-4o                                       & 82.92 & \textbf{100.00} & 95.25 & \textbf{100.00} & \underline{70.25} & \underline{76.33} & 72.42 \\ 
Gemini-1.5-pro                               & 66.50 & 99.42 & 99.50 & 72.92 & 41.00 & 23.08 & 53.00 \\
Claude-3-haiku                               & 67.25 & 99.33 & \underline{99.75} & 22.42 & 29.25 & 32.08 & 4.83 \\ 
Claude-3-opus                                & 65.58 & 99.67 & 99.58 & \textbf{100.00} & 54.08 & 56.33 & 28.17 \\
Claude-3.5-sonnet                            & \underline{85.58} & 99.83 & \textbf{99.92} & \textbf{100.00} & 66.50 & 66.17 & \textbf{80.92} \\

\bottomrule                                                      
\end{tabular}
}
\caption{Performance comparison across LLMs on the summarization tasks. On the \textbf{Easy} tasks, both strong open-source and closed-source models achieve high performance. However, on the \textbf{Hard} split, DeepSeek-V3 and R1 exhibit superior instruction-following capabilities, outperforming all other models by a significant margin. Among closed-source models, GPT-4o-mini and GPT-4o show better performance compared to their counterparts.}
\label{table:04_summ_results}
\end{table}

When breaking down performance across the individual summarization tasks shown in \Cref{table:04_summ_results}, we notice that generating numbered points is easier for LLMs than bullet points, probably because generated numbers guide the LLM to stop at the correct number of points, satisfying the formatting constraint. 
Although all LLMs seemingly master producing numbered points, adding a constraint on the length of each point proves much harder: performance
is divided by a factor of 4 for many open-source LLMs (e.g. Mistral-7B, DS-Dis-Qwen-1.5B). Indenting points proves to be the hardest task and half the LLMs (including closed-source GPT-3.5-turbo and Claude-3-haiku) collapse to near-zero accuracy. DeepSeek-R1 shows a strong lead on all LLMs for all three Hard summarization tasks.

\begin{figure*}
    \centering
    \includegraphics[width=\linewidth]{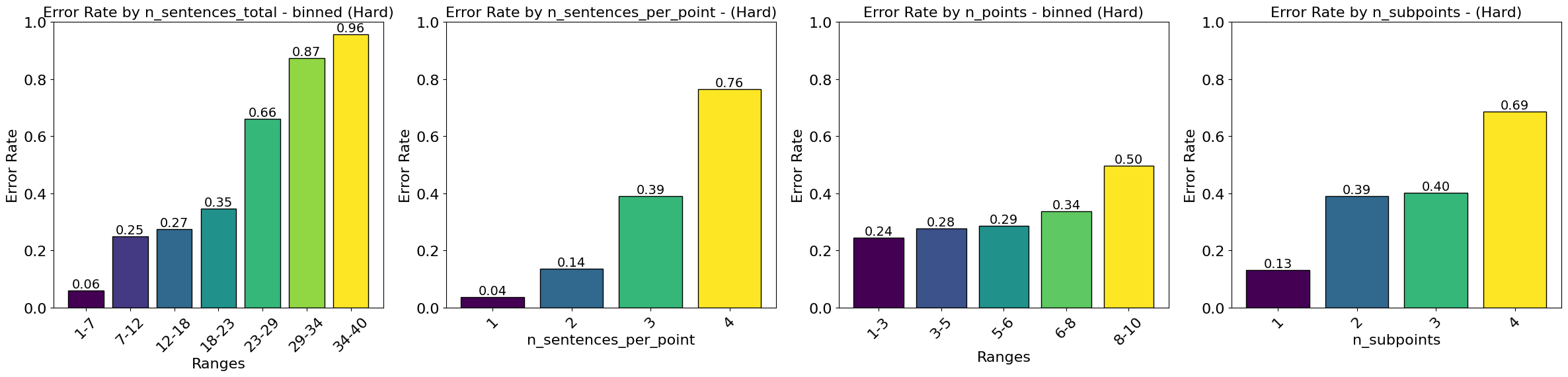}
    \caption{Error rate of GPT-4o across different features of the Summarization Bullet Points+Length (Hard) task. As the number of required key points increases or specific organizational requirements are added, the error rate rises significantly.}
    \label{fig:binned_summ}
\end{figure*}

A further analysis with GPT-4o in \Cref{fig:binned_summ} shows error rate for binned values of the Hard formatting condition \emph{Bullet points + length} (controlling the length of each bullet point). Length control error rate jumps beyond 20 total sentences, or 4 sentences per point. This finding proves that longer outputs are hard to structure and format for LLMs.

\paragraph{Code Results}

As we compare the models in \Cref{table:04_main_results}, we observe that DeepSeek-v3 achieves the best performance among open-source models, which can be attributed to its larger parameter size and higher-quality pre-training. DeepSeek-R1 also delivers competitive results, closely followed by Llama-3.1-70B. Among the closed-source models, Claude-3.5-sonnet and Claude-3-opus stand out as the top performers.

\begin{table}[t]
\centering
\resizebox{0.9\linewidth}{!}{
\begin{tabular}{lcccccccc}
\toprule

\multirow{2}{*}{\textbf{LLM}} & \multicolumn{2}{c}{\centering\textbf{Add Print}} & \multicolumn{2}{c}{\textbf{Replace Vars}} & \multicolumn{2}{c}{\textbf{Input Gen}} & \multicolumn{2}{c}{\textbf{Simulate Exec}} \\ \cline{2-9} 
                              & \textbf{Easy}           & \textbf{Hard}          & \textbf{Easy}       & \textbf{Hard}       & \textbf{Easy}      & \textbf{Hard}     & \textbf{Easy}        & \textbf{Hard}       \\ \hline
DS-Dis-Qwen-1.5B              & 32.50                   & 0.0                    & 32.08               & 7.29                & 6.67               & 0.00              & 49.93                & 9.55                \\
Phi-3-mini-128k               & 70.42                   & 0.50                   & 84.58               & 43.28               & 12.50              & 0.00              & 38.75                & 9.55                \\
Qwen-2-7B                     & 60.83                   & 1.01                   & 79.17               & 41.00               & 22.92              & 0.00              & 39.58                & 11.05               \\
Mistral-7B                    & 47.08                   & 0.50                   & 25.00               & 17.31               & 33.75              & 0.00              & 19.17                & 6.03                \\
Llama-3.1-8B                  & 77.50                   & 2.01                   & 85.83               & 53.08               & 1.66               & 3.52              & 36.67                & 10.55               \\
Mistral-nemo                  & 74.17                   & 2.51                   & 82.08               & 52.16               & 50.42              & 0.00              & 45.83                & 16.58               \\
Mixtral-8x7B                  & 40.42                   & 0.50                   & 12.50               & 9.34                & 40.83              & 0.50              & 39.58                & 5.53               \\
Llama-3.1-70B                 & \underline{95.00}       & 21.61                  & 87.92               & 64.92               & \underline{66.67}  & \underline{4.02}  & 71.25                & 22.61      \\
DeepSeek-v3                   & \textbf{96.25}          & \textbf{24.12}         & \underline{88.33}   & \textbf{69.48}      & \textbf{73.75}     & \textbf{5.53}     & \textbf{91.25}       & \underline{34.67}     \\
DeepSeek-R1                   & \underline{95.00}       & \underline{22.11}      & \textbf{89.58}      & \underline{66.52}   & 50.00              & 2.01              & \underline{84.17}    & \textbf{57.79}      \\
\hdashline GPT-3.5-turbo      & 76.25                   & 0.00                   & 90.42               & 57.40               & \underline{72.92}  & 1.51              & 58.33                & 18.59               \\
GPT-4o-mini                   & 90.00                   & 10.55                  & \underline{91.25}   & 66.51               & 66.25              & 3.02              & 82.08                & 22.61               \\
GPT-4o                        & 85.00                   & 9.55                   & 86.67               & 70.62               & \textbf{79.58}     & 4.52              & \textbf{94.17}       & \underline{32.66}   \\
Gemini-1.5-pro                & \underline{94.17}       & \underline{34.17}      & 83.33               & 70.62               & 65.83              & 4.02              & 85.42                & \textbf{43.22}      \\
Claude-3-haiku                & 75.42                   & 5.03                   & 86.67               & 60.59               & 40.00              & \underline{5.53}  & 62.92                & 17.59               \\
Claude-3-opus                 & \textbf{96.25}          & \textbf{40.20}         & \textbf{91.67}      & \textbf{78.82}      & 69.58              & 2.01              & 82.50                & 23.12               \\
Claude-3.5-sonnet             & 90.00                   & 9.55                   & \underline{91.25}   & \underline{78.59}   & 70.42              & \textbf{6.03}     & \underline{87.50}    & 24.62               \\ 
\bottomrule                                                      
\end{tabular}
}
\caption{Performance comparison across LLMs on code-related tasks shows that among open-source models, DeepSeek-v3 and R1 excel, particularly on the hard split. Among closed models, Gemini-1.5-pro, Claude-3-opus, and Claude-3.5-sonnet achieve higher accuracy.}
\label{tab:code-results}
\end{table}


From the perspective of the individual tasks shown in \Cref{tab:code-results}, we observe that Hard-level problems are significantly more complex, as longer code snippets increase the difficulty of comprehension. Additionally, tasks requiring deeper understanding and memorization present greater challenges. For instance, on the Easy level of \emph{Add Print Statements} and \emph{Replace Variables}, even smaller open-source models like Llama-3.1-8B achieve strong performance. However, only a few powerful closed-source models (e.g., the Claude series models, Gemini-1.5-pro, and GPT-4o ) perform well on the Hard level of \emph{Replace Variables}. Furthermore, nearly all models struggle with \emph{Test Case Inputs Generation} at the Hard level, as longer code snippets often involve multiple complex operations, such as loops, recursion, and switch-case statements.
On the Hard split \emph{Simulate Execution}, we find most smaller models fail to pass more than 20\% questions. Among the close-sourced models, Gemini-1.5-pro demonstrates significantly better performance but is also far behind DeepSeek-R1.


\paragraph{HTML Results}

From the results in Table \ref{table:04_main_results}, we observe that open-source models generally underperform compared to closed-source models in both Easy and Hard HTML generation tasks, with higher accuracies in the Easy task than in the Hard task. Among open-source models, DeepSeek-v3 leads, followed by DeepSeek-R1, while among closed-source models, Claude-3.5-sonnet is the top performer, closely trailed by Claude-3-opus and GPT-4o. Notably, DeepSeek-v3, DeepSeek-R1, Claude-3.5-sonnet, Claude-3-opus, and GPT-4o also rank among the highest in MMLU scores (Table \ref{table:04_correlation_results}). Additionally, larger models generally outperform smaller ones, as seen with Llama-3.1-70B compared to Llama-3.1-8B.

\begin{figure}[t]
  \centering
  \begin{minipage}[t]{0.55\textwidth}
    \centering
    \includegraphics[width=\linewidth]{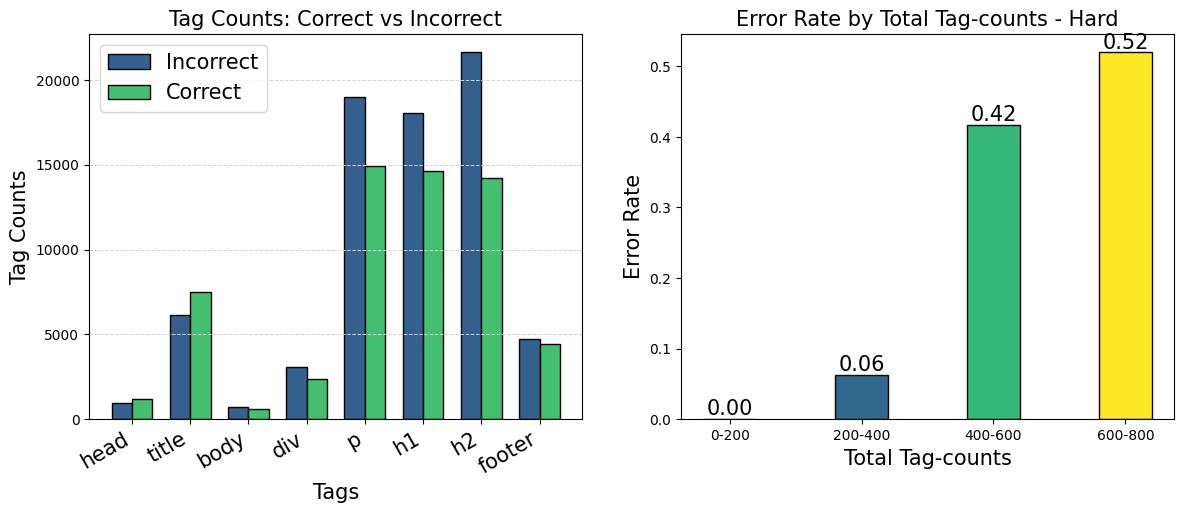}
    \caption{\small Tag-counts for correct vs. incorrect HTML generations (left) and error rate by total tag counts (binned) (right) for the Hard task in GPT-4o.}
    \label{fig:html_error_rate}
  \end{minipage}
  \hfill 
  \begin{minipage}[t]{0.4\textwidth}
    \centering
    \includegraphics[width=\linewidth]{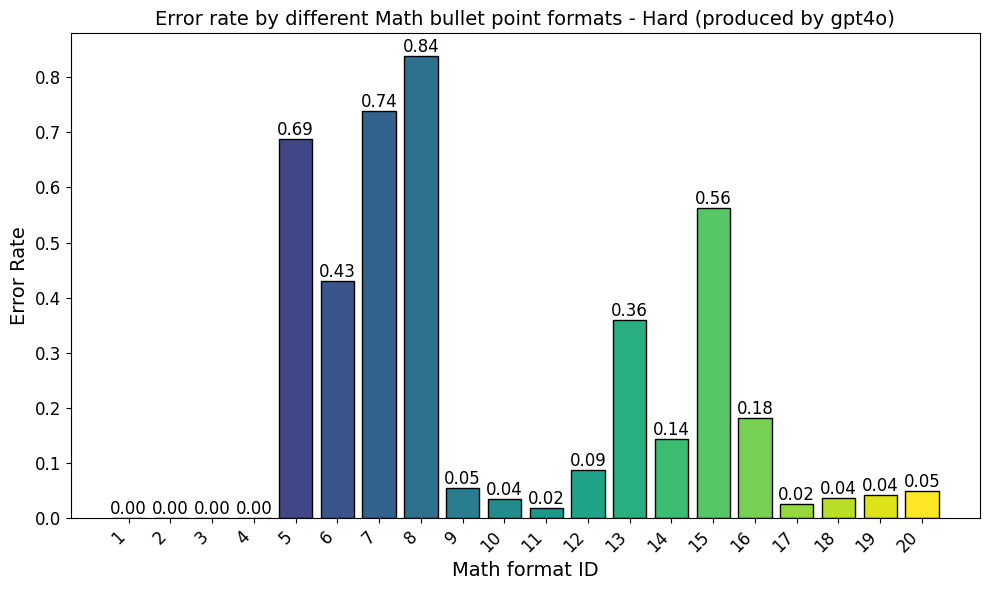}
    \caption{\small Error rates of GPT-4o in GSM8K math reasoning across 20 Hard formats.}
    \label{img:math_error_rate}
  \end{minipage}
\end{figure}

We further provide two analyses based on GPT-4o's performance on the Hard task in \Cref{fig:html_error_rate}: one examines the distribution of cumulative tag-counts for each tag (Section \ref{subsec:Structured-HTML}) in both correct and incorrect HTML code generation samples, and the other analyzes the distribution of all tag-counts in incorrect HTML code generation samples. Both figures reveal a consistent trend of increasing error rates as the number of tag-counts grows, confirming that LLMs struggle with structured HTML code generation, particularly when tasked with producing a larger number of HTML tags. This issue is especially pronounced for deeply nested tags like div'', p'', h1'', and h2'', as these tags are generated multiple times more frequently than their parent containers due to their nesting structure.



\paragraph{Math Results}


In Table \ref{table:04_main_results} we present the math format-following percentage accuracy for the Easy (final answer style) and Hard (final answer and bullet point style) categories, using GSM8K as the underlying benchmark. To be considered correct, an answer must be both {accurate} and {\textbf{compliant with the corresponding format requirement}}. Consistently, most models perform significantly worse in both Easy and Hard settings compared to their scores in standard benchmarks \citep{llm-eval-harness}. For instance, Gemini-1.5-pro achieves 77.33\% in Easy and 73.39\% in Hard settings, while originally scoring 91.7\% in the standardized test. In fact, while most closed-source models in Table \ref{table:04_main_results} exceed 90\% in standard benchmarks \citep{llm-eval-harness}, they experience significant performance drops in our math evaluations, with margins as high as 70\%. This indicates that these models are not as reliably or consistently proficient in math as previously assumed and may have overfitted to specific formats and styles. Notably, smaller and older closed-source models like GPT-3.5-turbo and Claude-3-haiku show considerable degradation, with scores below 40\%. Similarly, most open-source models, such as Mixtral-8x7B, perform even worse, dropping below 10\% accuracy on the Hard split. However, DeepSeek-R1 demonstrates stronger resilience in maintaining format compliance, likely due to its {deep reasoning process}. Overall, these results suggest that existing math reasoning comparisons between models are likely unreliable and unfair unless tested across a wide variety of diverse and impartial formats. Our framework offers a more robust alternative for such evaluations.

To provide deeper insights, Figure \ref{img:math_error_rate} illustrates the error rates of GPT-4o on GSM8K when tested across 20 Hard formats. Despite being a highly advanced frontier model, GPT-4o exhibits widely varying performance depending on the format. Specifically, it achieves perfect scores (zero error rate) in formats 1 to 4 but struggles in others, with error rates as high as 84\%. This suggests that the model may have overfitted to certain popular formats while faltering with novel ones. Further manual inspection reveals that the model often produces accurate final answers but fails to adhere to the instructed formats, resulting in these samples being marked as incorrect.



\subsection{Correlation to General Reasoning} 
\label{sec:correlation}

\begin{figure}[h!]
    \centering
    \begin{minipage}[t]{0.50\textwidth}
        \centering
        \vspace{0pt}
        \resizebox{0.99\linewidth}{!}{
        \begin{tabular}{lccc}
        \toprule
        \multicolumn{1}{l}{\textbf{LLM}} & \multicolumn{1}{c}{\textbf{StructTest}} & \multicolumn{1}{c}{\textbf{Arena}} & \multicolumn{1}{c}{\textbf{MMLU}} \\
        \toprule
        Phi-3-mini-128k   & 18.81                          & 1,037                         & 68.10                    \\
        Mistral-7B        & 14.08                          & 1,072                         & 60.10                     \\
        Llama-3.1-8B      & 37.22                          & 1,175                         & 73.00                    \\
        Mixtral-8x7B      & 18.12                          & 1,114                         & 70.60                    \\
        Llama-3.1-70B     & 65.69                          & 1,248                         & 86.00                    \\
        DeepSeek-V3       & 73.66                          & 1,319                         & 88.50           \\
        DeepSeek-R1       & 72.15                          & 1,361                         & 90.80           \\
        \hdashline
        GPT-3.5-turbo     & 38.27                          & 1,068                         & 70.00                    \\
        GPT-4o-mini       & 60.04                          & 1,272                         & 82.00                    \\
        GPT-4o            & 73.50                          & 1,265                         & 88.70                    \\
        Gemini-1.5-pro    & 63.44                          & 1,302                         & 85.90                     \\
        Claude-3-haiku    & 36.15                          & 1,179                         & 75.20                    \\
        Claude-3-opus     & 68.81                          & 1,247                         & 86.80                    \\
        Claude-3.5-sonnet & 72.62                          & 1,268                         & 88.70          \\
        \bottomrule
        \end{tabular}
        }
        \captionof{table}{Comparison of StructTest average accuracy with ChatBot Arena scores and MMLU accuracy. ChatBot Arena results are current as of March 13th, 2025.
        }
        \label{table:04_correlation_results}    
    \end{minipage}
    \hfill 
    \begin{minipage}[t]{0.48\textwidth}
        \centering
        \raisebox{-\height}{
        \includegraphics[width=\linewidth]{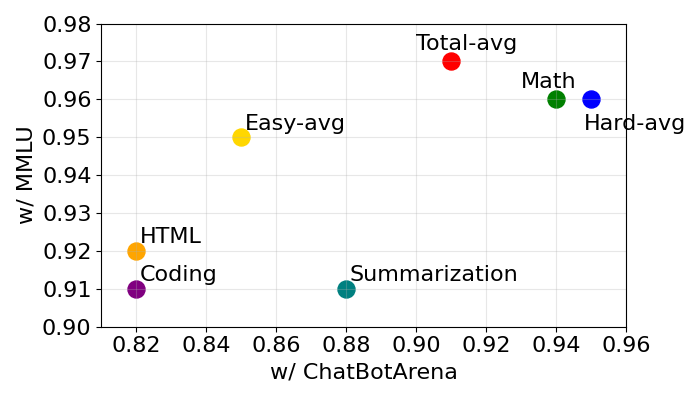}
        }
        \captionof{figure}{Correlation of various StructTest setups with ChatBot Arena and MMLU. {The Hard and Total splits show strong correlation with both MMLU and ChatBot Arena, with math demonstrating the highest correlation among the four domains.}}
        \label{fig:correlation_plot}
    \end{minipage}
\end{figure}

To understand whether StructTest could be a ``cheap'' proxy of general reasoning ability in LLMs, we compare the average accuracy in StructTest with popular benchmarks including ChatBot Arena and MMLU in
\Cref{table:04_correlation_results}. We include all the models for which we could find both Arena and MMLU scores. The correlation (Pearson's product-moment coefficient) between StructTest and Arena is \textbf{92.5}\% and the same for StructTest and MMLU is \textbf{96.3}\%. Such results highlight that StructTest, though being naturally unbiased, cheap to evaluate, and robust to data contamination, offers strongly correlated evaluation results compared to expensive and resource-intensive benchmarks like ChatBot Arena and MMLU.

To better understand how different task domains and difficulty settings influence correlation with existing benchmarks, we present a 2D scatter plot in \Cref{fig:correlation_plot}. 
As shown in the figure, all domains and the Hard split exhibit high correlation with MMLU and Chatbot Arena. 
This suggest that StructTest can effectively serve as proxy for general reasoning without potential risk in data contamination and bias as it can be actively and easily extended.



\subsection{Discussion about Update StructTest On-the-fly}

With the rapid advancement of LLMs, it is increasingly important to update existing benchmarks dynamically to maintain their reliability for newly released models and mitigate potential data contamination. From its inception, StructTest has been designed with extendability and scalability in mind, though achieving this goal will also require collaborative efforts from the broader community. Specifically, updates can be made in three key areas:

\textbf{Rules~~} Despite extensive efforts to scrutinize model predictions, some cases may still arise where models exploit loopholes to artificially inflate scores. When such cases are identified, the corresponding evaluation rules can be updated to enhance robustness.

\textbf{Data~~} As discussed earlier, most tasks in StructTest do not rely on annotated benchmark data, and they are decoupled from underlying data by rule-based evaluation. This enables the seamless integration of newly collected raw corpora, further minimizing the risk of data contamination.

\textbf{Tasks~~} Similar to rule updates, new tasks can be introduced by carefully designing evaluation rules. This aspect relies more on community contributions. When failures are identified in powerful models, these patterns can be analyzed to develop new tasks for StructTest, ensuring its continued relevance and challenge.




\section{Conclusion}
We have proposed StructTest, a programmatically verifiable benchmark for evaluating instruction-following capabilities of LLMs through structured outputs. 
StructTest is a cheap-to-run and unbiased benchmark with adjustable difficulty levels, which is especially robust to the prevailing issue of contamination among existing LLM benchmarks. 
By conducting evaluation across 17 popular LLMs, we find that it remains challenging even for the very best models like DeepSeek-R1, GPT-4o or Claude-3.5-Sonnet, which all score below 70\% accuracy on the Hard subset of StructTest.
Notably, the open-source DeepSeek-R1 shines in StructTest as its performance is comparable to the top closed-source models. 
Besides, lower results on the math domain compared with those on the standardized benchmarks reveal the potential overfitting to answer format of existing LLMs.
Our analysis of correlation with other benchmarks (i.e., MMLU and ChatBot Arena) shows that StructTest serves as a good proxy for evaluating general reasoning ability in LLMs. We believe that  StructTest offers a critical, complementary approach to existing LLM evaluations. 




\bibliography{colm2025_conference,custom}
\bibliographystyle{colm2025_conference}

\appendix

\section{Closed-Source Model Versions} 
\label{appendix:model_versions}

We show the API version used in our evaluation results for close-source models in \Cref{tab:api_version}. The inference for all closed-source models was performed during 27th November 2024 to 14th December 2024.
\begin{table}[h]
\centering
\resizebox{0.6\linewidth}{!}{
\begin{tabular}{ll}
\toprule
\textbf{Model}    & \textbf{API Version}       \\
\toprule
GPT-3.5-turbo     & gpt-3.5-turbo-0125         \\
GPT-4o-mini       & gpt-4o-mini-2024-07-18     \\
GPT-4o            & gpt-4o-2024-08-06          \\
Gemini-1.5-pro    & gemini-1.5-pro-002         \\
Claude-3-haiku    & claude-3-haiku-20240307    \\
Claude-3-opus     & claude-3-opus-20240229     \\
Claude-3.5-sonnet & claude-3-5-sonnet-20241022
\\ \bottomrule
\end{tabular}
}
\label{tab:api_version}
\caption{Closed-source model versions used in Evaluation Results}
\end{table}

\section{Examples for Different Tasks}
\label{appendix:task_examples}

\begin{figure*}
    \centering
    \includegraphics[width=0.92\textwidth]{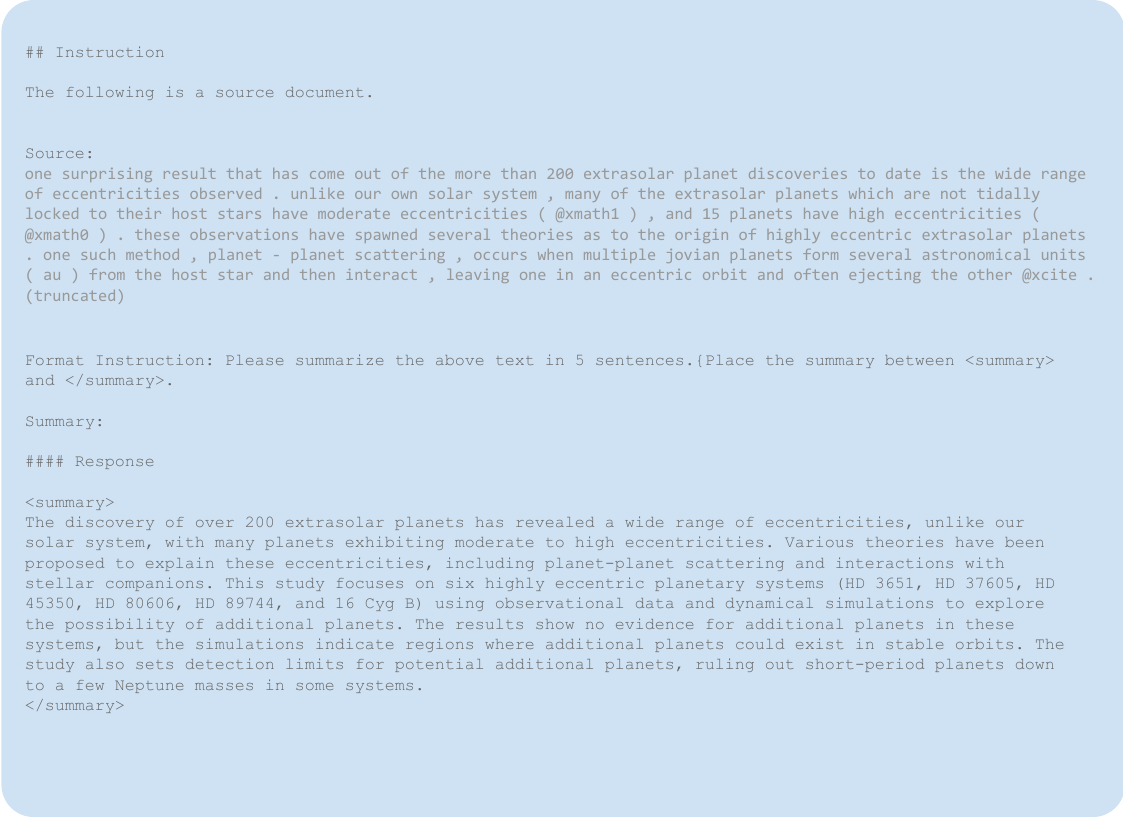}
    \caption{Test example for \textbf{length} task in Summarization.}
    \label{fig:length-summ}
\end{figure*}

\begin{figure*}
    \centering
    \includegraphics[width=0.92\textwidth]{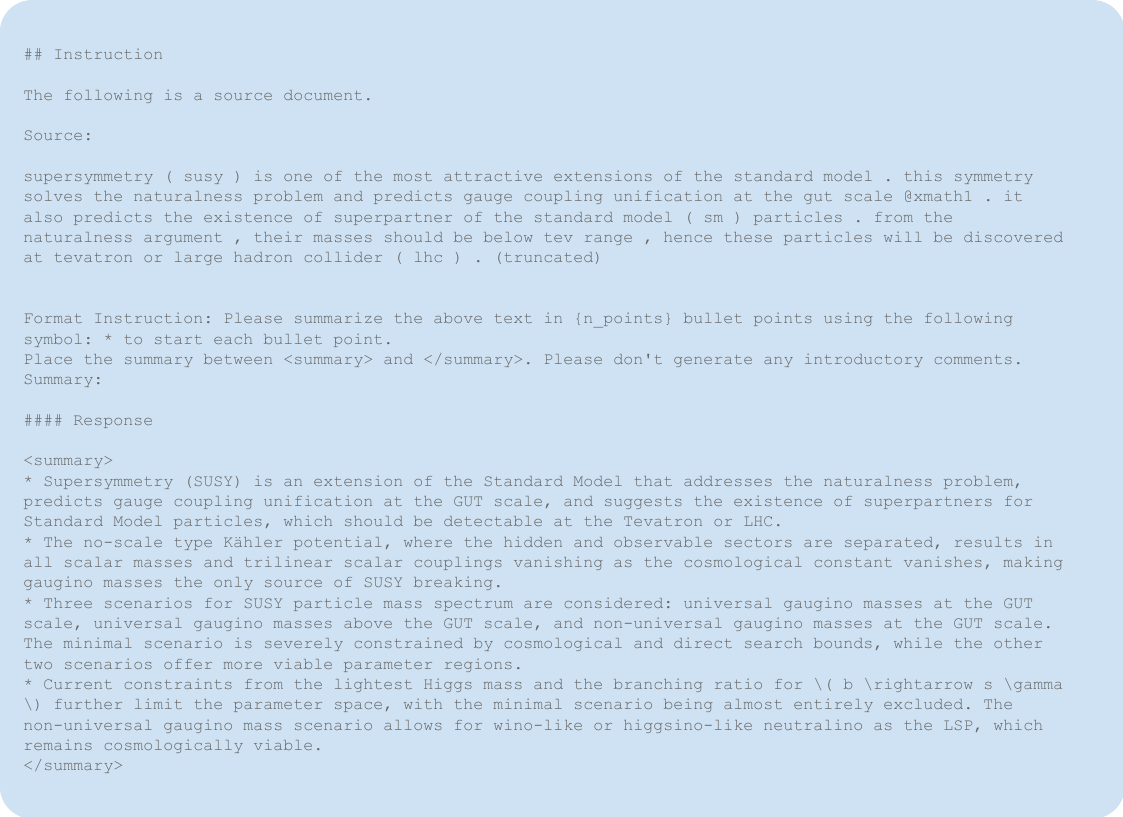}
    \caption{Test example for \textbf{bullet points} task in Summarization.}
    \label{fig:bullet-points-summ}
\end{figure*}

\begin{figure*}
    \centering
    \includegraphics[width=0.92\textwidth]{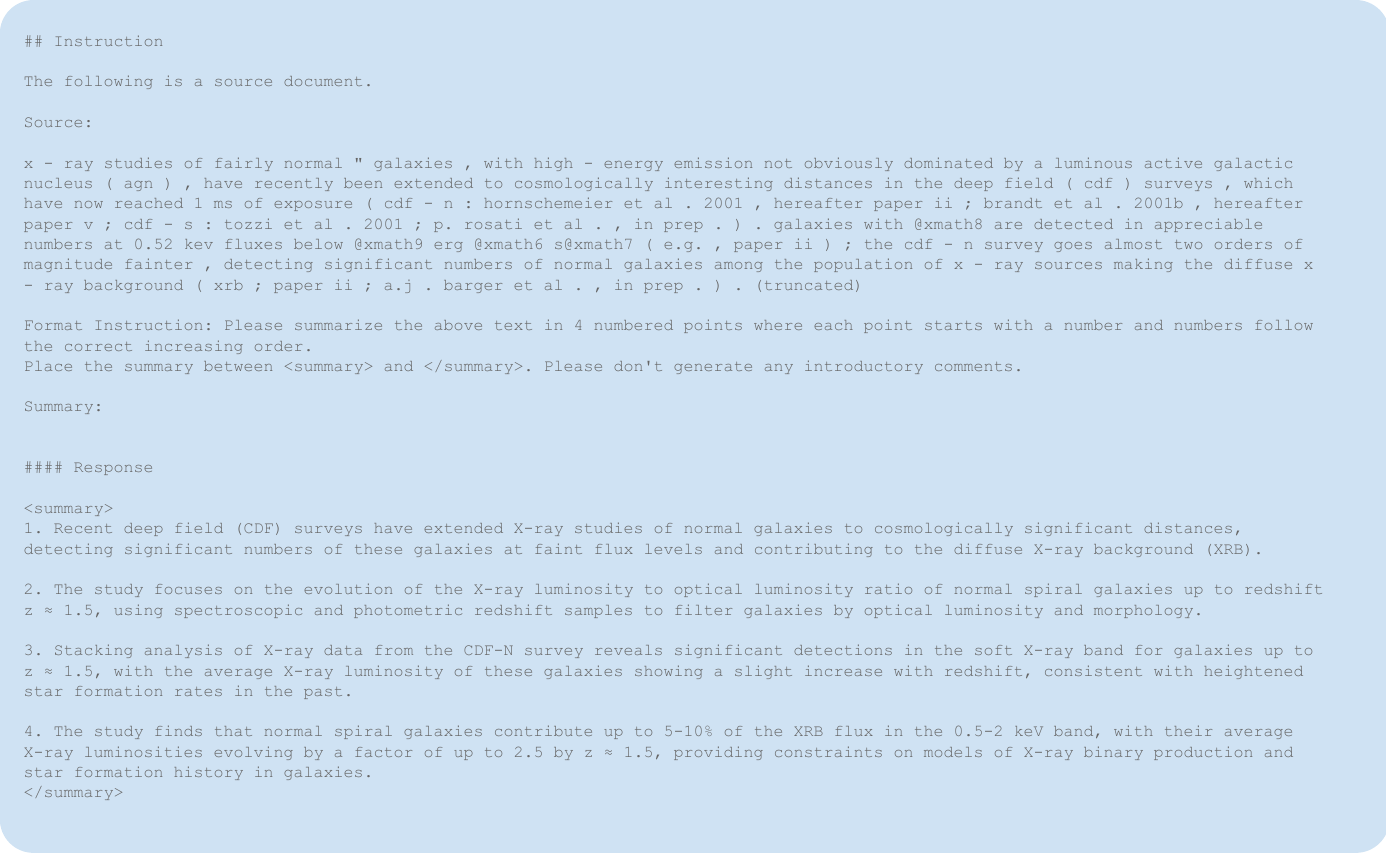}
    \caption{Test example for \textbf{numbered points} task in Summarization.}
    \label{fig:numbered-points-summ}
\end{figure*}

\begin{figure*}
    \centering
    \includegraphics[width=0.92\textwidth]{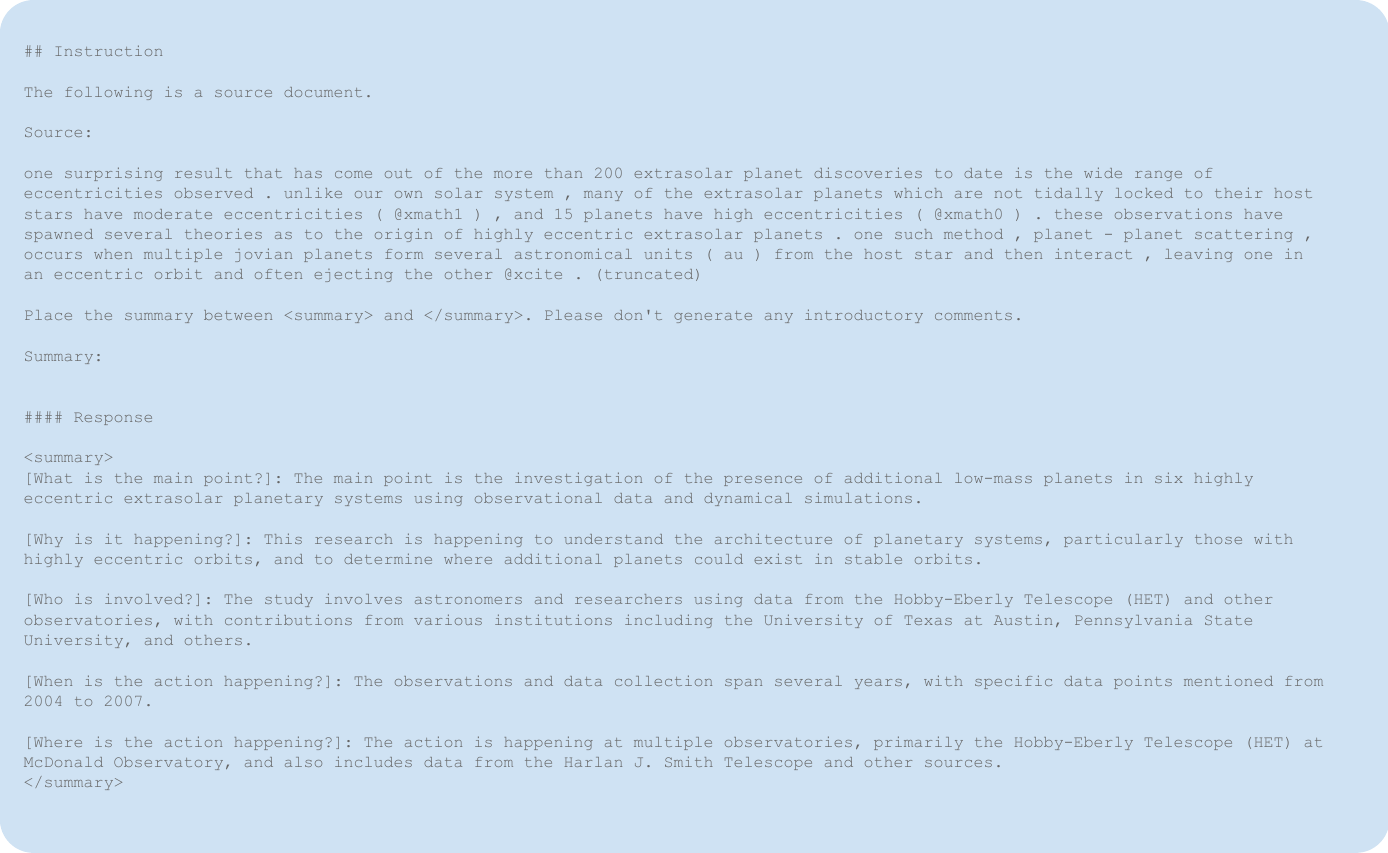}
    \caption{Test example for \textbf{questions} task in Summarization.}
    \label{fig:questions-summ}
\end{figure*}

\begin{figure*}
    \centering
    \includegraphics[width=0.92\textwidth]{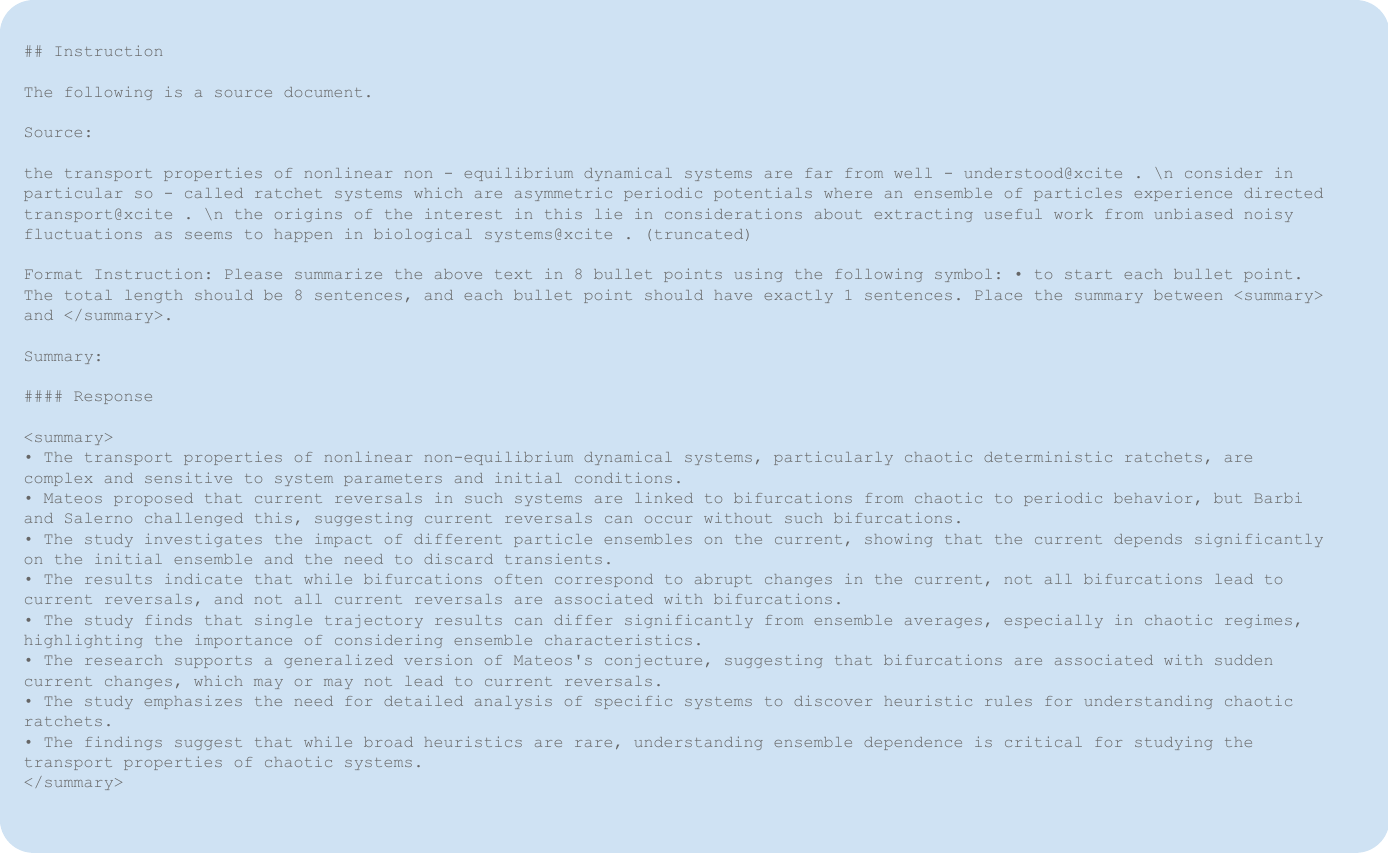}
    \caption{Test example for \textbf{combination of bullet points and length} task in Summarization.}
    \label{fig:bullet-points-length-summ}
\end{figure*}

\begin{figure*}
    \centering
    \includegraphics[width=0.92\textwidth]{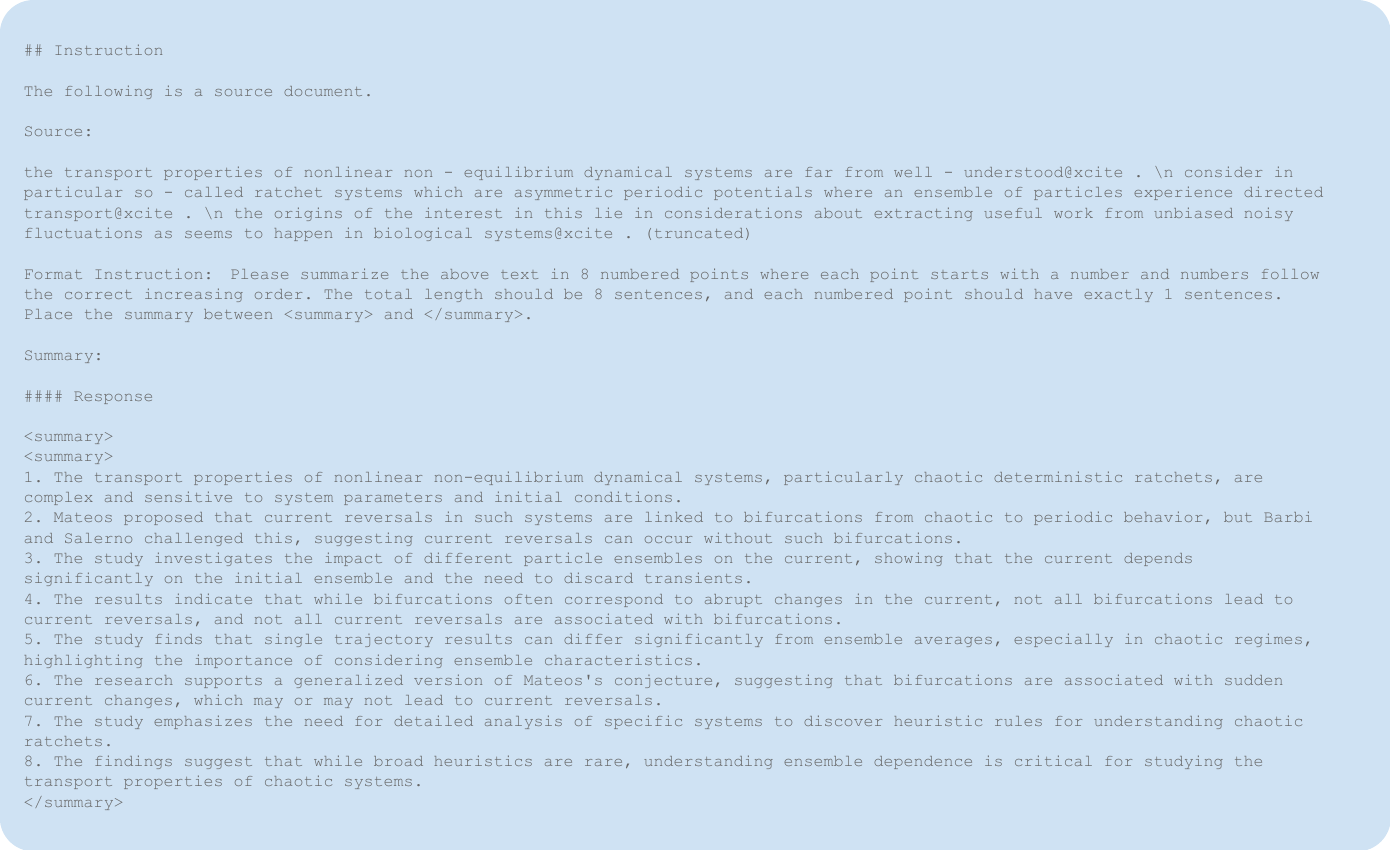}
    \caption{Test example for \textbf{combination of numbered points and length} task in Summarization.}
    \label{fig:numbered-points-length-summ}
\end{figure*}

\begin{figure*}
    \centering
    \includegraphics[width=0.92\textwidth]{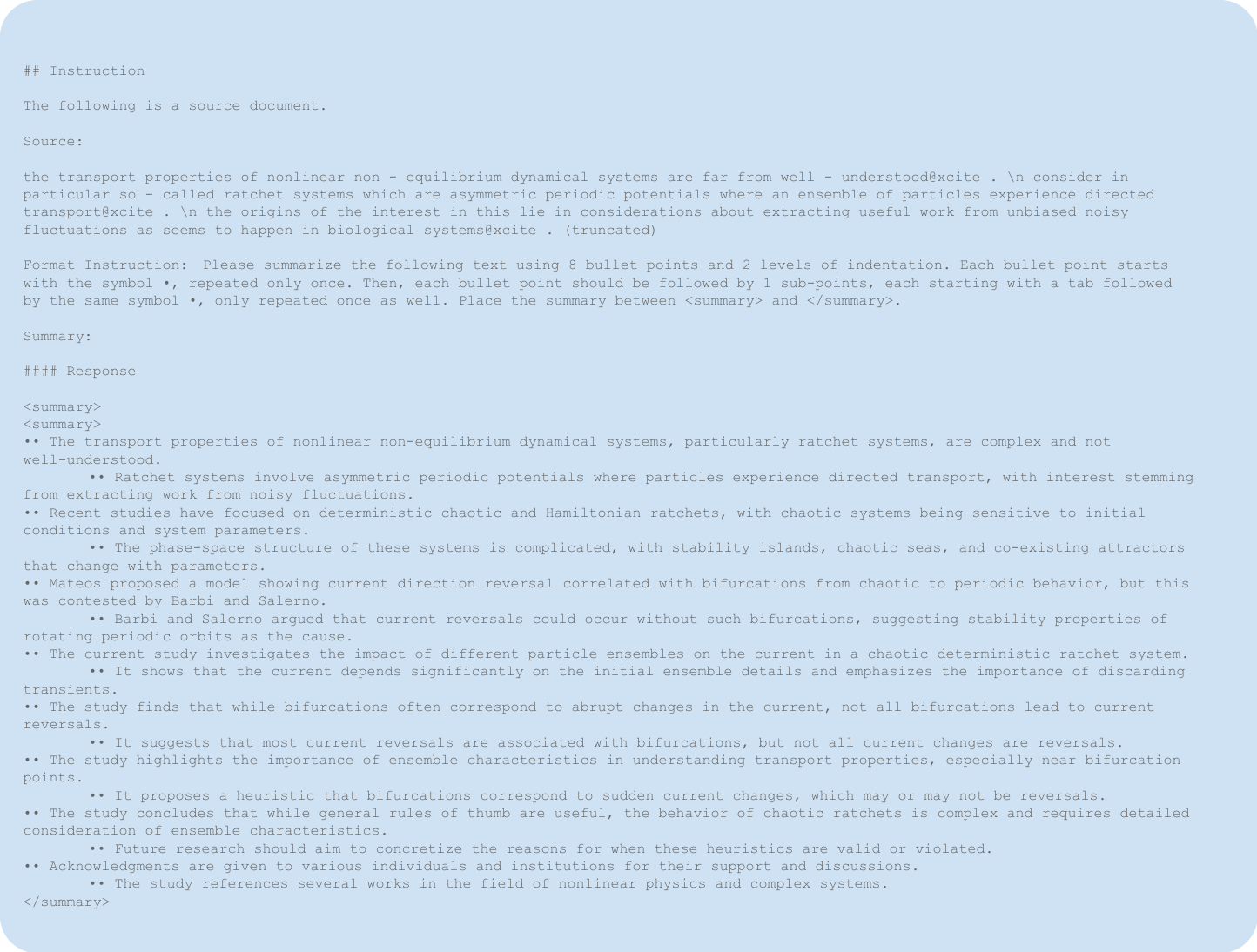}
    \caption{Test example for \textbf{indented bullet points} task in Summarization.}
    \label{fig:indented-bullet-points-summ}
\end{figure*}

\begin{figure*}
    \centering
    \includegraphics[width=0.92\textwidth]{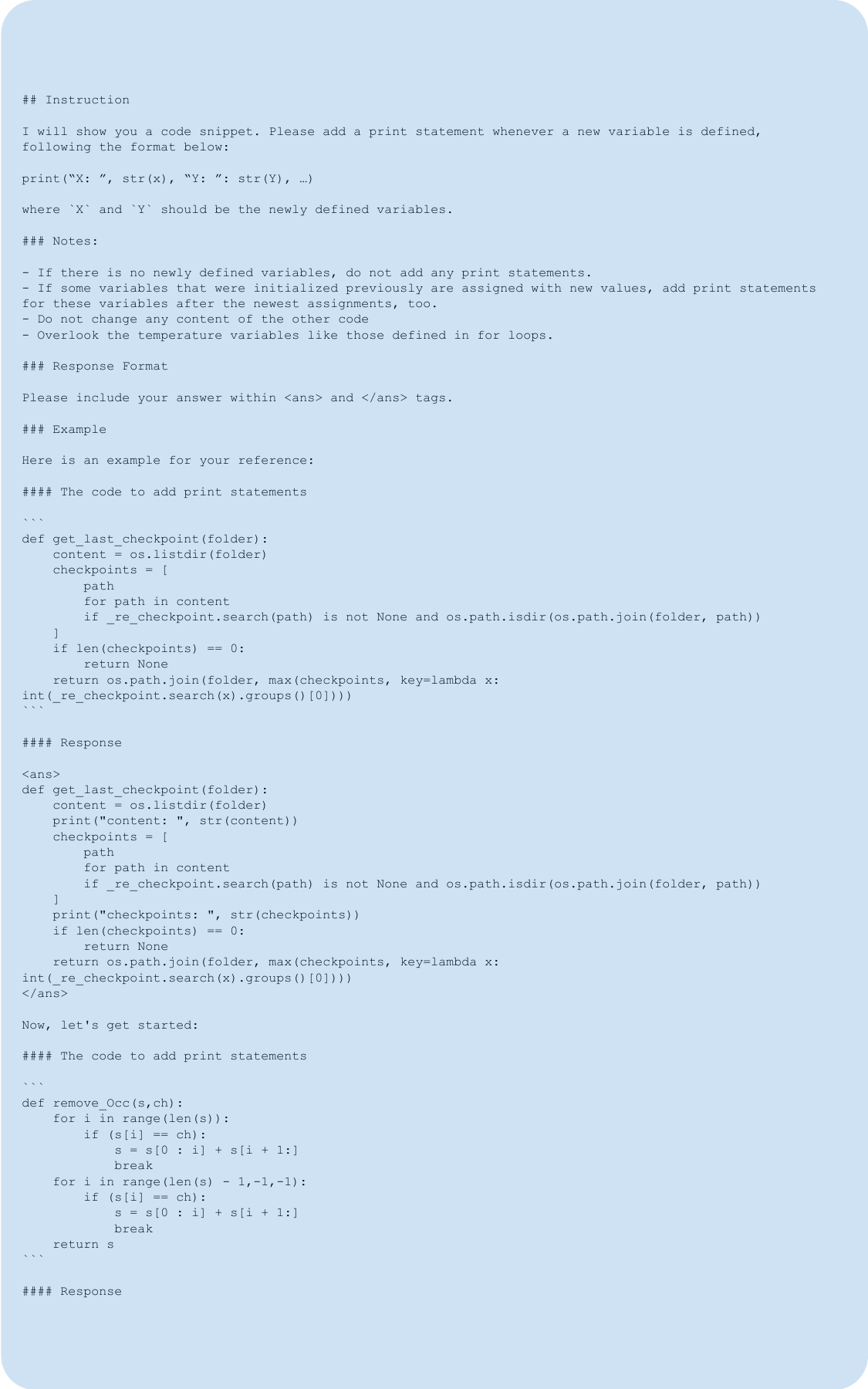}
    \caption{Test example for \textbf{add `print' statements} task with one-shot prompting.}
    \label{fig:exp-add-print-state}
\end{figure*}

\begin{figure*}
    \centering
    \includegraphics[width=0.92\textwidth]{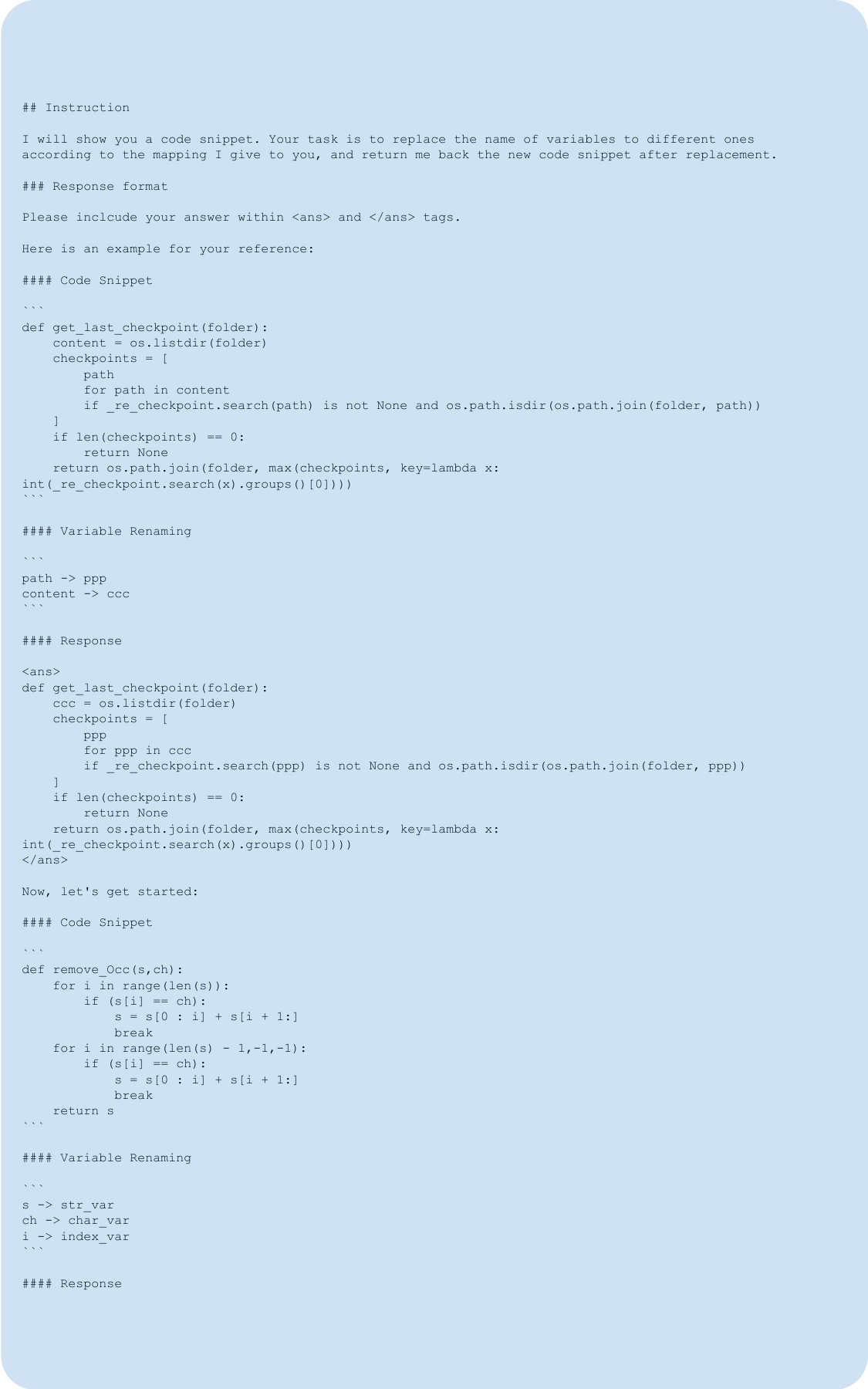}
    \caption{Test example for \textbf{replace variables} task with one-shot prompting.}
    \label{fig:exp-replace-variable}
\end{figure*}

\begin{figure*}
    \centering
    \includegraphics[width=0.92\textwidth]{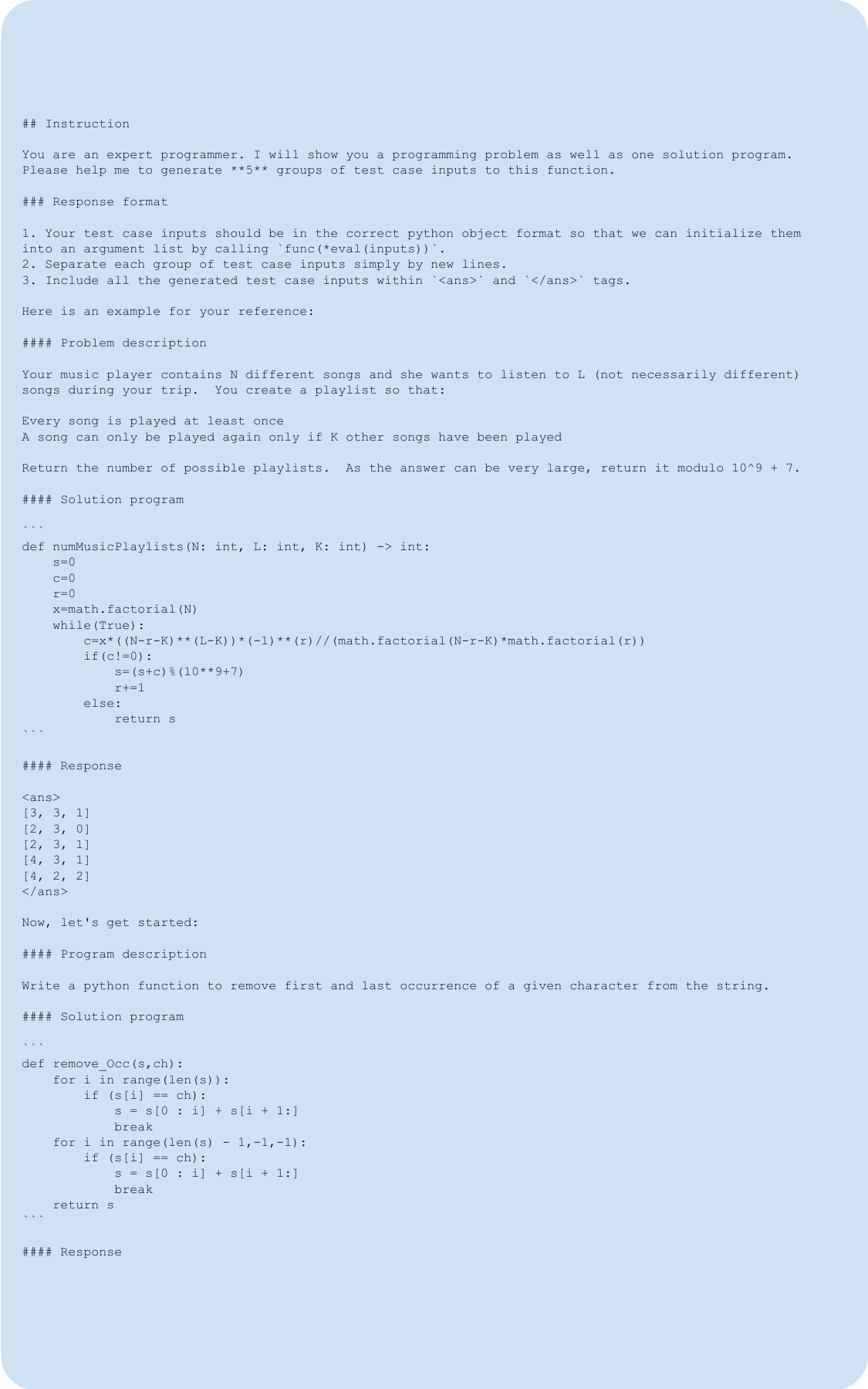}
    \caption{Test example for \textbf{test case input generation (easy)} task with one-shot prompting. Easy level task contains only function-based code, whereas the hard level task is mainly composed of problems using standard input-output stream.}
    \label{fig:exp-test-case-inputs-gen}
\end{figure*}

\begin{figure*}
    \centering
    \includegraphics[width=0.92\textwidth]{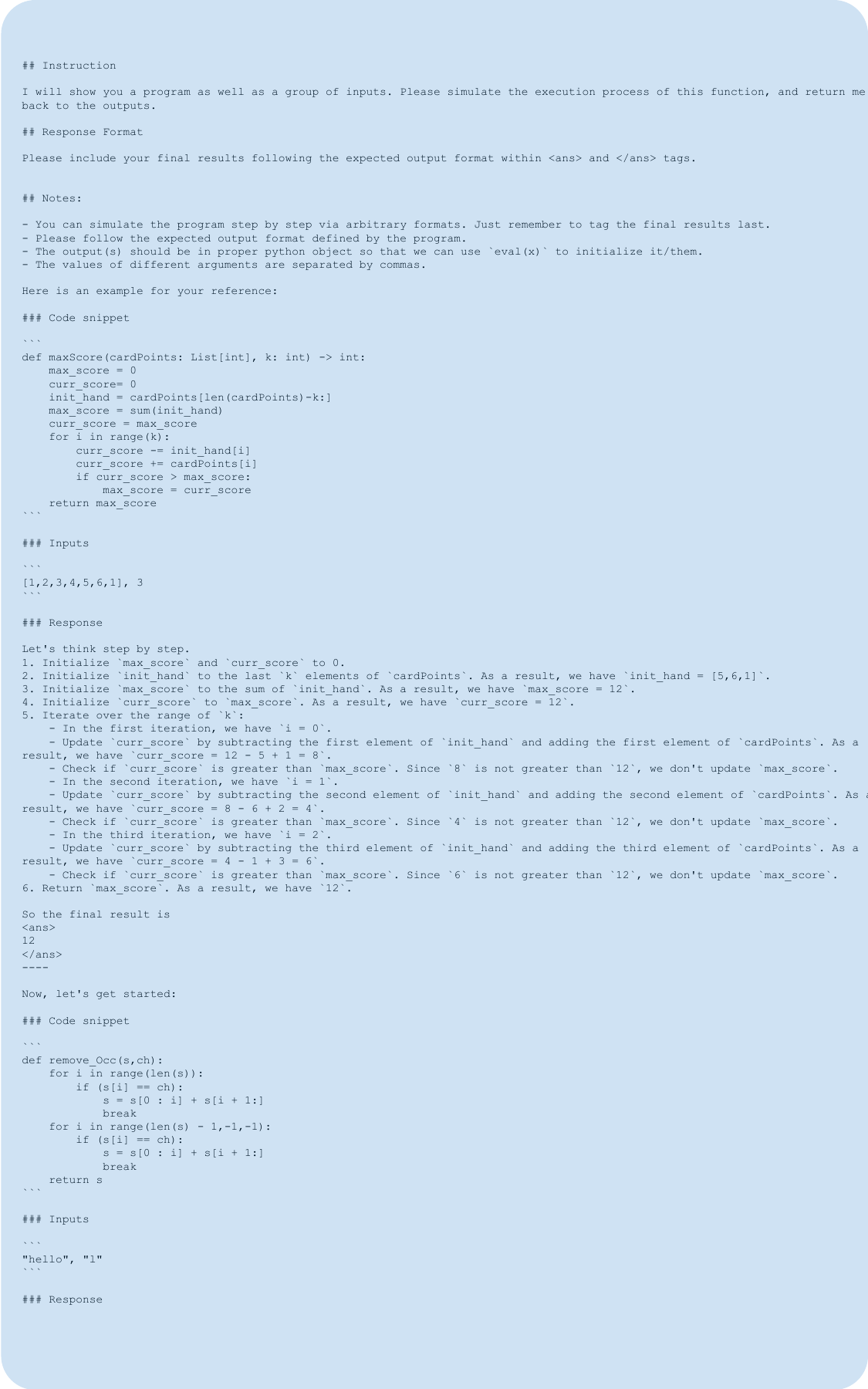}
    \caption{Test example for \textbf{simulate execution} task with one-shot prompting.}
    \label{fig:exp-simulate-exec}
\end{figure*}

We show examples for each summarization task in \Cref{fig:length-summ,fig:bullet-points-summ,fig:numbered-points-summ,fig:questions-summ,fig:bullet-points-length-summ,fig:numbered-points-length-summ,fig:indented-bullet-points-summ}, and examples for each coding task in \Cref{fig:exp-add-print-state,fig:exp-replace-variable,fig:exp-test-case-inputs-gen,fig:exp-simulate-exec}.

\begin{figure*}
    \centering
    \includegraphics[width=0.92\textwidth]{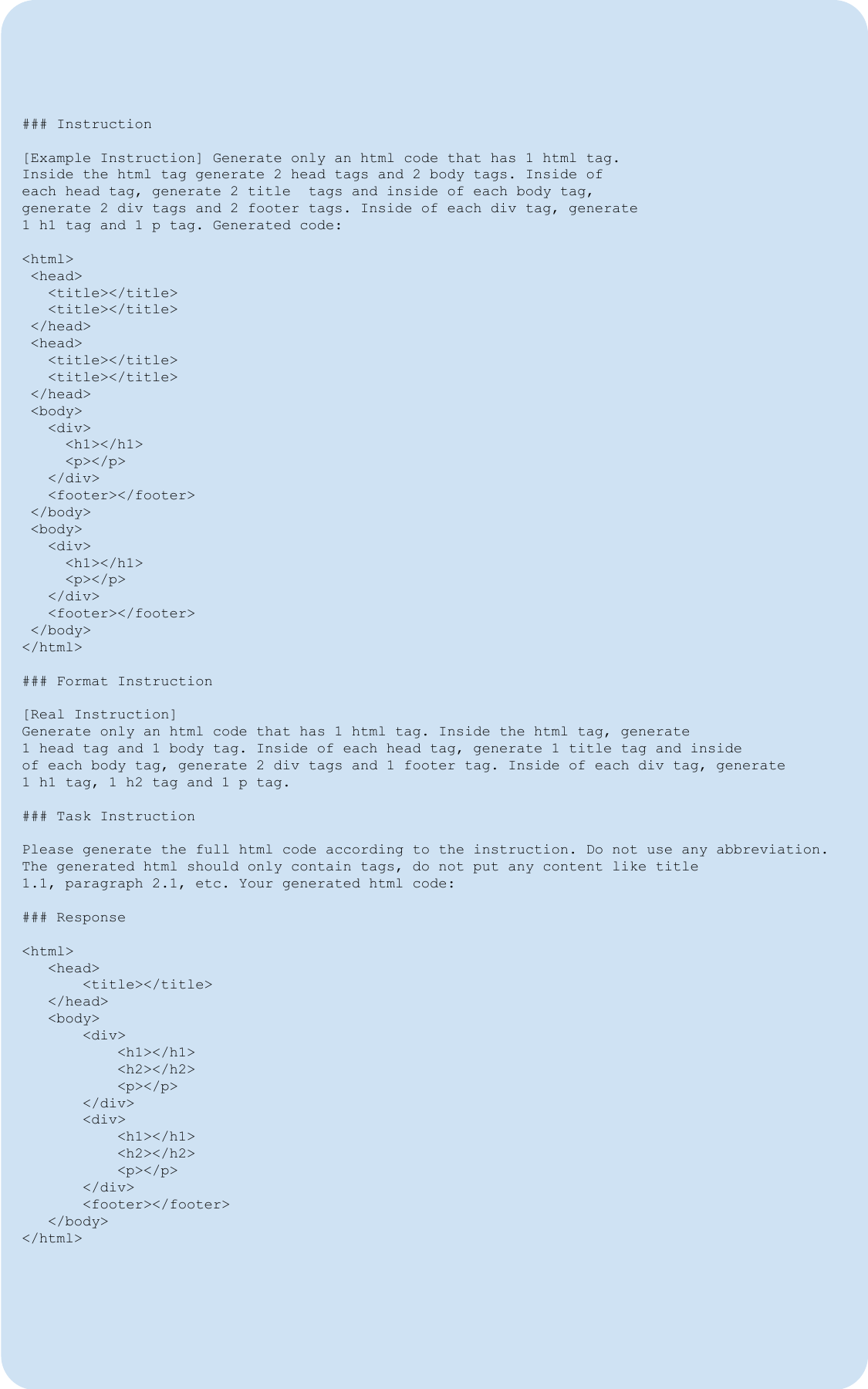}
    \caption{Test example for \textbf{html generation} task with one-shot prompting.}
    \label{fig:exp-html-gen}
\end{figure*}

\label{sec:appendix}

\end{document}